\documentclass[11pt,a4paper]{article}
\usepackage[hyperref]{emnlp2020}
\usepackage{times}
\usepackage{latexsym}
\usepackage{multirow}
\usepackage{subfig}

\usepackage{amsmath}
\usepackage{amssymb}
\usepackage{graphicx}
\usepackage{booktabs}
\usepackage{multirow}

\usepackage{microtype}

\aclfinalcopy 

\newcommand{\ourmodel}[0]{SAFER}

\title{Graph-based Modeling of Online Communities for Fake News Detection}

\author{Shantanu Chandra$^\clubsuit$,\,\, Pushkar Mishra$^\bigstar$,\,\, Helen Yannakoudakis$^\spadesuit$,\,\, Madhav Nimishakavi$^\bigstar$,\,\,\\ \textbf{Marzieh Saeidi$^\bigstar$,\,\, Ekaterina Shutova$^\clubsuit$}\\
    $^\clubsuit$ ILLC, University of Amsterdam, The Netherlands \\
  $^\bigstar$ Facebook AI, London, United Kingdom\\
  $^\spadesuit$ Dept. of Informatics, King's College London, United Kingdom\\
  \normalsize{{\tt shanchandra93@yahoo.in, helen.yannakoudakis@kcl.ac.uk, e.shutova@uva.nl,}}\\
  \normalsize{{\tt \{pushkarmishra, madhavn, marzieh \}@fb.com}}
}
\date{}

\begin{document}
\maketitle

\begin{abstract}
Over the past few years, there has been a substantial effort towards automated detection of fake news on social media platforms. Existing research has modeled the structure, style, content, and patterns in dissemination of online posts, as well as the demographic traits of users who interact with them. However, no attention has been directed towards modeling the properties of online communities that interact with the posts. In this work, we propose a novel social context-aware fake news detection framework, \textbf{\ourmodel}, based on graph neural networks (GNNs). The proposed framework aggregates information with respect to: 1) the nature of the content disseminated, 2) content-sharing behavior of users, and 3) the social network of those users. We furthermore perform a systematic comparison of several GNN models for this task and introduce novel methods based on relational and hyperbolic GNNs, which have not been previously used for user or community modeling within NLP.
We empirically demonstrate that our framework yields significant improvements over existing text-based techniques and achieves state-of-the-art results on fake news datasets from two different domains.
\end{abstract}


\section{Introduction}
The spread of fake news online leads to undesirable consequences in many areas of societal life, notably in the political arena and healthcare with the most recent example being the COVID-19 ``Infodemic'' \cite{Infodemic}. Its consequences include political inefficacy, polarization of society and alienation among individuals with high exposure to fake news \cite{fakenews_effects1,fakenews_effects2}. Recent years have therefore seen a growing interest in automated methods for fake news detection, which is typically set up as a binary classification task. While a large proportion of work has focused on modeling the structure, style and content of a news article \cite{conv_han, style_based_auto_detect}, no attempts have been made to understand and exploit the online community that interacts with the article.

To advance this line of research, we propose \textbf{\ourmodel{}} (\textbf{S}ocially \textbf{A}ware \textbf{F}ake n\textbf{E}ws detection f\textbf{R}amework), a graph-based approach to fake news detection that aggregates information from 1) the content of the article, 2) content-sharing behavior of users who shared the article, and 3) the social network of those users. We frame the task as a graph-based modeling problem over a heterogeneous graph of users and the articles shared by them. We perform a systematic comparison of several graph neural network (GNN) models as graph encoders in our proposed framework and introduce novel methods based on relational and hyperbolic GNNs, which have not been previously used for user or community modeling within NLP. By using relational GNNs, we explicitly model the different relations that exist between the nodes of the heterogeneous graph, which the traditional GNNs are not designed to capture. Furthermore, euclidean embeddings used by the traditional GNNs have a high distortion when embedding real world hierarchical and scale-free graphs\footnote{A Scale Free Network is one in which the distribution of links to nodes follows a power law, i.e., the vast majority of nodes have very few connections, while a few important nodes (hubs) have a huge number of connections.} \cite{eucl_distort1, eucl_distort2}. Thus, by using hyperbolic GNNs we capture the relative distance between the node representations more precisely by operating in the hyperbolic space. Our methods generate rich community-based representations for articles. We demonstrate that, when used alongside text-based representations of articles, \ourmodel{} leads to significant gains over existing methods for fake news detection and achieves state-of-the-art performance. We also make the code publicly available\footnote{\url{https://github.com/shaanchandra/SAFER}}.


\section{Related Work}
Approaches to fake news detection can be categorized into three different types: \textit{content}-, \textit{propagation}- and \textit{social-context} based. Content-based approaches model the content of articles, such as the headline, body text, images and external URLs. 
Some methods utilize knowledge graphs and subject-predicate-object triples \cite{spo_1,spo_2}, while other feature-based methods model writing style, psycho-linguistic properties of text, rhetorical relations and content readability \cite{lang_style, tfidf_style_based, style_based_auto_detect,  style_based_stylometric}. Others use neural networks \cite{simple_rnn_fakenews}, with attention-based architectures such as HAN \cite{han_fakenews} and dEFEND \cite{defend} outperforming other neural methods. Recent multi-modal approaches encoding both textual and visual features of news articles as well as tweets \cite{multi_modal_1, eann}, have advanced the performance further.

Propagation-based methods analyze patterns in the spread of news based on \textit{news cascades} 
\cite{fakenews_survey} which are tree structures that capture the content's post and re-post patterns. These methods make predictions in two ways: 1) computing the similarity between the cascades \cite{graph_kernels_orig2, cascade_similarity}; or 2) representing news cascades in a latent space for classification \cite{recursive_nn}. However, they are not well-suited to large social-network setting due to their computational complexity.

Social-context based methods employ the users' meta-information obtained from their social media profiles (e.g. geo-location, total words in profile description, etc.) as features for detecting fake news \cite{social_cntxtt_1, social_cntxt_2}. Recently, several works have leveraged GNNs to learn user representations for other tasks, such as abuse \cite{abuse_detection}, political perspective \cite{pol_persp_gcn} and stance detection \cite{gat_stance}. 

Two works, contemporaneous to ours, have also proposed to use GNNs for the task of fake news detection. \citet{gnn_continual_learning} applied GNNs on a \textit{homogeneous} graph constructed in the form of news cascades by using just shallow user-level features such as no. of followers, status and tweet mentions. 
On the other hand, \citet{fang} use features derived from the article, news source, users and their interactions and timeline of posting
to detect fake news. They construct two \textit{homogeneous} sub-graphs (\textit{news-source} and \textit{user} sub-graph) 
and model them separately in an unsupervised setting for proximity relations. They also use the user's stance in relation to the shared content as additional information via a stance detection network pre-trained on a self-curated dataset. 

Our formulation of the problem is distinct from these methods in three ways. Firstly, we construct a single \textit{heterogeneous} graph consisting of two kinds of nodes and edges and model them \textit{together} in a semi-supervised graph learning setup. Secondly, we do not perform user profiling, but rather compute community-wide social-context features, and to the best of our knowledge, no prior work has investigated the role of online communities in fake news detection. 
Third, to capture the role of communities, we only use the information about the users' networks, without the need for any personal information from user's profile and yet outperform the existing methods that incorporate those.
Furthermore, since our methods do not use any user-specific information, such as their location, race or gender, they therefore do not learn to associate specific population groups with specific online behaviour, unlike other methods that explicitly incorporate user-specific features and their personal information by design. We believe the latter would pose an ethical concern, which our techniques help to alleviate.



\begin{figure*}[t!]
    \centering
        \includegraphics[width=0.95\textwidth]{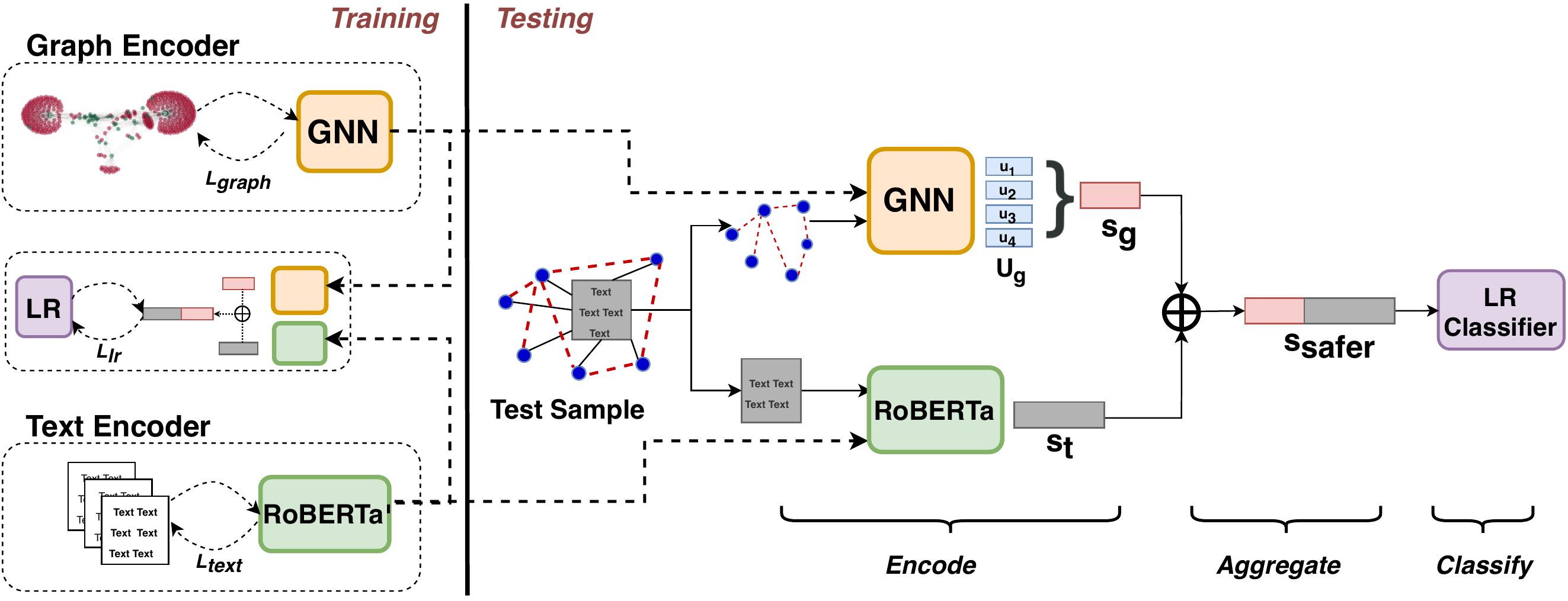}
    \caption{Visual representation of the proposed \ourmodel{} framework. Graph and text encoders are trained independently followed by training of a logistic regression (LR) classifier. During inference, the text of the article as well as information about its social network of users are encoded by the trained text and graph encoders respectively. Finally, the social-context and textual features of the article are concatenated for classification using the trained LR classifier.}
    \label{fig:safer}
\end{figure*}

\section{Datasets}
For our experiments, we use fake news datasets from two different domains, i.e., celebrity gossip and healthcare, to show that our proposed method is domain-agnostic. All user information collected for the experiments is de-identified.

\vspace{1mm}
\noindent
\textbf{FakeNewsNet\footnote{\url{https://tinyurl.com/uwadu5m}}} \cite{fakenewsnet} is a
a publicly available benchmark for fake news detection. The dataset contains news articles from two fact-checking sources, \textit{PolitiFact} and \textit{GossipCop}, along with links to Twitter posts mentioning these articles. \textit{PolitiFact}\footnote{\url{https://www.politifact.com/}} 
is a fact-checking website for political statements; \textit{GossipCop}\footnote{\url{https://www.gossipcop.com/}} 
is a website that fact-checks celebrity and entertainment stories. \textit{GossipCop} contains a substantially larger set of articles compared to \textit{PolitiFact} (over $21$k news articles with text vs. around $900$) and is therefore the one we use in our experiments. We note that some articles have become unavailable over time. We also excluded $60$ articles that are less than $25$ tokens long. In total, we work with $20,350$ articles of the original set ($23$\% fake and $77$\% real). 

\vspace{1mm}
\noindent
\textbf{FakeHealth\footnote{\url{https://tinyurl.com/y36h42zu}}} \cite{fakehealth}, is a publicly available benchmark for fake news detection specifically in the healthcare domain. The dataset is collected from healthcare information review website \textit{Health News Review}\footnote{\url{https://www.healthnewsreview.org/}}, which reviews whether a news article is reliable according to 10 criteria and gives it a score from 1-5. In line with the original authors of the dataset, we consider an article as fake for scores less than 3 and real otherwise. The dataset is divided into two datasets based on the nature of the source of the articles. \textit{HealthStory} contains articles that are news stories, i.e., reported by news media such as Reuters Health. \textit{HealthRelease} contains articles that are news releases from various institutions such as universities, research centers and companies. \textit{HealthStory} contains a considerably larger set of articles compared to \textit{HealthRelease} (over 1600 vs around 600) and is therefore the one we use in our experiments. We again note that some articles have become unavailable over time. We also exclude $27$ articles that are less than $25$ tokens long. In total, we work with $1611$ of the original set ($28$\% fake and $72$\% real articles).


\section{Methodology}

\subsection{Constructing the Community Graph}
\label{sec:graph_construct}
For each of the datasets, we create a heterogeneous community graph $\mathcal{G}$ consisting of two sets of nodes: user nodes $\mathcal{N}_u$ and article nodes $\mathcal{N}_a$. An article node $a \in \mathcal{N}_a$ is represented by a binary bag-of-words (BOW) vector $a = [w_{1}, .., w_{j}, .., w_{|V|}]$, where $\mathcal{|V|}$ is the vocabulary size and $w_i \in \{0,1\}$. A user node $u \in \mathcal{N}_u$ is represented by a binary BOW vector constructed over all the articles that they have shared: $u = [a_{1} \, | \, a_{2} \, | \, ... \, | \, a_{M}]$, where $|$ denotes the element-wise logical \textsc{or} and $M$ is the total number of articles shared by the user.
Next, we add undirected edges of two types: 1) between a user and an article node if the user shared the article in a tweet/retweet (article nodes may therefore be connected to multiple user nodes), and 2) between two user nodes if there is a follower--following relationship between them on Twitter.\footnote{We use Twitter APIs 
to retrieve the required information.} We work with the ``top N most active users" (N=20K for HealthStory, N=30K for GossipCop) subset and motivate this decision in Section \ref{sec:optim_support}. To avoid effects of any bias from frequent users, we exclude users who have shared more than $30$\% of the articles in either class. The resulting graph has $29,962$ user nodes, 
$16,766$ article nodes (articles from test set excluded) and over $1.2M$ edges for GossipCop. Meanwhile, the \textit{HealthStory} community graph contains $12,266$ user nodes, $1291$ article nodes (test set articles excluded) and over $450K$ edges.

\subsection{\ourmodel{}: Fake news detection framework}

The proposed framework -- detailed below and visualized in Figure \ref{fig:safer} -- employs two components in its architecture, namely graph- and text-based encoders, and its working can be broken down into two phases: training and testing. 

\vspace{2mm}
\noindent
\textbf{Training phase:} 
We first train the graph and text encoders independently on the training set. The input to the text encoder is the text of the article and it is trained on the task of article classification for fake news detection. The trained text encoder generates the text-based features of the article content $s_t \in \mathbb{R}^{d_t}$ where $d_t$ is the hidden dimension of the text encoder. 
The graph encoder is a GNN that takes as input the community graph (constructed as detailed in \S\ref{sec:graph_construct}). The GNN is trained with supervised loss from the article nodes that is back-propagated to the rest of the network. The trained GNN is able to generate a set of user embeddings $U_g = \{u_1, u_2, ..., u_m\}$ where $u_i \in \mathbb{R}^{d_g}$, $d_g$ is the hidden dimension of the graph encoder and $m$ is the total number of users that interacted with the article. These individual user representations are then aggregated into a single fixed size vector via normalized sum such that $s_g = \big(\sum_{i=1}^m u_i \big)/m, s_g \in \mathbb{R}^{d_g}$ where $s_g$ denotes the social-context features of the article. The final social context-aware representation of the article is computed as $s_{safer} = s_g \oplus s_t$, where $\oplus$ is the concatenation operator. This form of aggregation helps \ourmodel{} to retain the information that each representation encodes about different aspects of the shared content. 
Finally, $s_{safer}$ is used to train a logistic regression (LR) classifier on the training set. 
Intuitively, the trained text encoder captures the linguistic cues from the content that are crucial for the task. Similarly, the trained graph encoder learns to assign users to implicit online communities based on their content-sharing patterns and social connections. 

\vspace{2mm}
\noindent
\textbf{Testing phase:} To classify unseen content as fake or real, \ourmodel{} takes as input the text of the article as well as the network of users that interacted with it. It then follows the same procedure as detailed above to generate the social context-aware representation of the to-be-verified test article, $s_{safer}$, and uses the trained LR classifier to classify it.

\subsection{Text Encoders}

We experiment with two different architectures as text encoders in \ourmodel{}:

\vspace{1mm}
\noindent
\textbf{Convolutional neural network (CNN)}. We adopt the sentence-level encoder of \citet{kim-cnn} at the document level. This model uses multiple 1-D convolution filters of different sizes that aggregate information by sliding over the length of the article. The final fixed-length article representation is obtained via max-over-time pooling over the feature maps.  

\vspace{1mm}
\noindent
\textbf{RoBERTa}. As our main text encoder, we fine-tune the transformer encoder architecture, RoBERTa \cite{roberta}, and use it for article classification. RoBERTa is a language model pre-trained with dynamic masking. Specifically, we use it to encode the first $512$ tokens of each article and use the \texttt{\textsc{[cls]}} token as the article embedding for classification.

\subsection{Graph Encoders}
We experiment with six different GNN architectures for generating user embeddings as detailed below: 

\vspace{1mm}
\noindent
\textbf{Graph Convolution Networks (GCNs).} GCNs \cite{gcn} 
take as input a graph $\mathcal{G}$ defined by its  adjacency matrix $A \in \mathbb{R}^{n \times n}$ (where $n$ is number of nodes in the graph)\footnote{$A$ is symmetric, i.e., $A_{ij} = A_{ji}$, with self-loops $A_{ii} = 1$.}, a degree matrix $D$ such that $D_{ii} = \sum_j A_{ij}$, and a feature matrix $F \in \mathbb{R}^{n \times m}$ containing the $m$-dimensional feature vectors for the nodes. The recursive propagation step of a GCN at the $i^{th}$ convolutional layer is given by: $O^i = \sigma(\tilde{A} O^{(i-1)} W^i)$
where $\sigma$ denotes an activation function, $\tilde{A} = D^{-\frac{1}{2}} A D^{-\frac{1}{2}}$ is the degree-normalized adjacency matrix; $W^i \in \mathbb{R}^{t_{i-1} \times t_i}$ is the weight matrix of the $i^{th}$ convolutional layer; $O^{(i-1)} \in \mathbb{R}^{n \times t_{i-1}}$ represents the output of the preceding convolution layer and $t_i$ is the number of hidden units in the $i^{th}$ layer, with $t_0 = m$. 

\vspace{1mm}
\noindent
\textbf{Graph Attention Networks (GAT).} GAT \cite{gat} is a non-spectral architecture that leverages the spatial information of a node directly by learning different weights for different nodes in a neighborhood using a self-attention mechanism. 
GAT is composed of graph attention layers. In each layer, a shared, learnable linear transformation $W \in \mathbb{R}^{t_{i-1} \times t_i}$ is applied to the input features of every node, where $t_i$ is the number of hidden units in layer $i$. Next, self-attention is applied on nodes, where a shared attention mechanism 
computes attention coefficients $e_{uv}$ between pairs of nodes to indicate the importance of the features of node $v$ to node $u$.
To inject graph structural information, masked attention is applied by computing $e_{uv}$ only for nodes $v \in \mathcal{U}(u)$ that are in the first-order neighborhood of node $u$.
The final node representation is obtained by linearly combining normalized attention coefficients with their corresponding neighborhood node features.

\vspace{1mm}
\noindent
\textbf{GraphSAGE}. SAGE \cite{graphsage} is an inductive framework that learns aggregator functions that generate node embeddings from a node's local neighborhood.
First, each node $u \in \mathcal{G}$ aggregates information (through either mean, LSTM or performing max-pooling after passing them through a linear layer) from its local neighborhood $h_{v}^{k-1}, \forall v \in \mathcal{U}(u)$ into a single vector $h_{\mathcal{U}(u)}^{k-1}$ where $k$ denotes the depth of the search, $h^k$ denotes the node's representation at that step and $\mathcal{U}(u)$ is set of neighbor nodes of $u$. Next, it concatenates the node's current representation $h_{u}^{k-1}$ with that of its aggregated neighborhood vector $h_{\mathcal{U}(u)}^{k-1}$. This vector is then passed through a multi-layer perceptron (MLP) with non-linearity to obtain the new node representation $h_u^k$ to be used at depth $k+1$. Once the aggregator weights are learned, the embedding of an unseen node can be generated from its features and neighborhood.

\vspace{1mm}
\noindent
\textbf{Relational GCN/GAT}. 
R-GCN \cite{rgcn} and R-GAT are an extension of GCN and GAT for relational data and build upon the traditional differentiable message passing framework. The networks accept input in the form of a graph $\mathcal{G} = (V,E,\mathcal{R})$ where $V$ denotes the set of nodes, $E$ denotes the set of edges connecting the nodes and $\mathcal{R}$ denotes the edge relations $(u, r, v) \in E$ where $r \in \mathcal{R}$ is a relation type and $u, v \in V$. The R-GCN forward pass update step is: 

\small
\begin{align*}
    h_{u}^{(i)}=\sigma\left(\left[\sum_{r \in \mathcal{R}} \sum_{l \in \mathcal{U}_{u}^{r}} \frac{1}{c_{u, r}} W_{r}^{(i-1)} h_{l}^{(i-1)}\right] +W_{0}^{(i-1)} h_{u}^{(i-1)}\right)
\end{align*}
\normalsize

\noindent where $h_{u}^{(i)}$ is the final node representation of node $u$ at layer $i$, $\mathcal{U}_u^r$ denotes the set of neighbor indices of node $u$ under relation $r \in \mathcal{R}$, $W_r$ is the relation-specific trainable weight parameter and $c_{u,r}$ is a task specific normalization constant that can either be learned or set in advance (such as $c_{u,r} = |\mathcal{U}_u^r|$). Note that each node's feature at layer $i$ is also informed of its features from layer $i-1$ by adding a self-loop to the data with a relation type learned using the trainable parameter $W_0$. Intuitively, this propagation step aggregates transformed feature vectors of first-order neighbor nodes through a normalized sum. R-GAT also follows the same setup, except the aggregation is done using the graph attention layer as described in GAT. This architecture helps us to aggregate information from user and article nodes selectively from our community graph. 


\vspace{1mm}
\noindent
\textbf{Hyperbolic GCN / GAT}. \citealt{hygcn} build upon previous work \cite{hnn, hnn2} to combine the expressiveness of GCN/GAT with hyperbolic geometry to learn improved representations for scale-free graphs. Hy-GCN/GAT first map the euclidean input to the hyperbolic space (we use the Poincaré ball model), which is the Riemannian manifold with constant negative sectional curvature -1/K. Next, analogous to the mean aggregation by the GCN, Hy-GCN computes the Fréchet mean \cite{frechet_mean} of a node's neighbours' embeddings while the Hy-GAT performs aggregation in tangent spaces using hyperbolic attention. Finally, Hy-GCN/GAT use hyperbolic non-linear activation function $\sigma^{\otimes^{K_{i-1}, K_i}}$ given the hyperbolic curvatures -1/$K_{i-1}$, -1/$K_{i}$ at layers $i-1$ and $i$ where $\otimes$ is the Möbius scalar multiplication operator. This is crucial as it allows the model to smoothly vary curvature at each layer.

\subsection{Baselines and Comparison Systems}
We compare the performance of the proposed framework with seven supervised classification methods: two purely text-based baselines, a user-sharing majority voting baseline, a GNN-based ``social baseline'' and three architectures from the literature.
\vspace{1mm}

\noindent
\textbf{Baselines}. The setup for the baselines is detailed below:

\textit{1. Text-baselines.} We use the CNN and RoBERTa architectures described earlier to obtain article representations. The input to the CNN encoder is \textsc{elm}o embeddings \cite{elmo} of the article tokens, while RoBERTa uses its own tokenizer to generate initial token representations.

\textit{2. Majority sharing baseline}. This simple baseline classifies articles as fake or real based on the sharing statistics of users that tweeted or retweeted about it. If, on average, the users that interact with an article have shared more fake articles, then the article is tagged as fake, and real otherwise.

\textit{3. Social Baseline}. We introduce a graph-based model that measures the effectiveness of purely structural aspects of the community graph captured by the GNNs (without access to text). The user node embeddings are constructed as described earlier, but with the article nodes being initialized randomly. Here, the community-based features solely capture properties of the network. The classification is done using just the social-context feature by an LR classifier.

\vspace{1mm}

\noindent
\textbf{Comparison Systems.} We compare the performance of the proposed framework with three methods from literature:

\textit{1. HAN} \cite{defend}. Hierarchical attention network first generates sentence embeddings using attention over (GRU-based) contextualised word vectors. An article embedding is then obtained in a similar manner by passing sentence vectors through a GRU and applying attention over the hidden states.

\textit{2. dEFEND} \cite{defend}. This method exploits contents of articles alongside comments from users. Comment embeddings are obtained from a single layer bi-GRU and article embeddings are generated using HAN. A cross-attention mechanism is applied over the two embeddings to exploit users' opinions and stance to better detect fake news.

\textit{3. SAFE} \cite{safe}: This method uses visual and textual features of the content. It uses a CNN to encode the textual as well as visual content of an article by initially processing the visual information using a pre-trained \textit{image2sentence} model\footnote{\url{https://tinyurl.com/y3s965o5}}. It then concatenates these representations to better detect fake news.





\section{Experiments and Results}

\textbf{Experimental setup.} We use $70\%$, $10\%$ and $20\%$ of the total articles as train, validation and test splits respectively for both datasets. For CNN we use 128 filters of sizes [3,4,5] each. For HAN and dEFEND we report the results in \citet{defend}, while for SAFE in \citet{safe}. We use the large version of \textsc{r}o\textsc{bert}a and fine-tune all layers. 
Due to class imbalance, we weight the loss from the fake class 3 times more (in line with the class frequency in each of the datasets) while optimizing the binary cross entropy loss of the sigmoid output from a 2-layer MLP in all our experiments. We use dropout \cite{dropout}, attention dropout and node masking \cite{mishra2020node} for regularization. We use 2-layer deep architectures for all the GNNs. For Hy-GCN/-GAT we train with learnable curvature. We run all experiments with $5$ random seeds using the AdamW \cite{adamw} optimizer (except for Hy-GCN/-GAT that use Riemannian Adam; \citealt{radam}) with an early stopping patience of $10$.
For GossipCop, we use a learning rate of $5 \cdot 10^{-3}$ for Hy-GCN/-GAT; $1 \cdot 10^{-4}$ for SAGE and R-GAT; $1 \cdot 10^{-3}$ for R-GCN; and $5 \cdot 10^{-4}$ for the rest. We use weight decay of $5 \cdot 10^{-1}$ for RoBERTA; $2 \cdot 10^{-3}$ for SAGE and R-GCN; $1 \cdot 10^{-3}$ for the rest. We use dropout of 0.4 for GAT and R-GCN; 0.2 for SAGE and R-GAT; 0.5 for CNN; and 0.1 for the rest. We use node masking probability of 0.1 for all the GNNs and attention dropout of 0.4 for RoBERTa. Finally, we use a hidden dimension of 128 for SAGE; 256 for GCN and Hy-GCN; and 512 for the rest.
Meanwhile for HealthStory, we use a learning rate of $1 \cdot 10^{-4}$ for SAGE; $1 \cdot 10^{-3}$ for R-GAT; $5 \cdot 10^{-3}$ GCN, Hy-GCN/GAT; and $5 \cdot 10^{-4}$ for the rest. We use weight decay of $5 \cdot 10^{-1}$ for RoBERTa; $2 \cdot 10^{-3}$ for GAT, SAGE and R-GCN; and $1 \cdot 10^{-3}$ for the rest. We use dropout of 0.4 for GCN; 0.1 for Hy-GCN/GAT and RoBERTa, 0.5 for CNN; and 0.2 for the rest. We use node masking probability of 0.2 for GAT and R-GCN; 0.3 for Hy-GCN/-GAT; and 0.1 for the rest. Finally, we use attention dropout of 0.4 for RoBERTa and a hidden dimension of 128 for SAGE; 256 for Hy-GAT and 512 for the rest.


\begin{table}[t!]
\footnotesize
\begin{centering}
    \begin{tabular}{@{}llcc@{}}
    \toprule
         & \textbf{Model} & \textbf{GossipCop} & \textbf{HealthStory} \\\midrule
        \multirow{5}{*}{\textbf{Text}} & \textsc{han}$^\dagger$ & 67.20 & - \\
         & d\textsc{efend}$^\dagger$ & 75.00 & - \\
         & \textsc{safe}$^\ddagger$ & 89.50 & - \\\cmidrule{2-4}
         & \textsc{cnn} & 66.73 &  53.81 \\
         &\textsc{r}o\textsc{bert}a & 68.55 & 57.54 \\\midrule
         & Maj. sharing baseline & 77.19 & 8.20 \\\midrule
        \multirow{16}{*}{\textbf{Graph}} & \textbf{Social baseline} & \\
         & \multicolumn{1}{l}{\textsc{sage}} & 87.11 & 43.05 \\
         & \multicolumn{1}{l}{\textsc{gcn}} & 88.37 & 44.86 \\
         & \multicolumn{1}{l}{\textsc{gat}} & 87.94 & 46.13 \\
         & \multicolumn{1}{l}{\textsc{r-gcn}} & 89.68 & 46.28 \\
         & \multicolumn{1}{l}{\textsc{r-gat}} & 89.21 & 46.89 \\
         & \multicolumn{1}{l}{\textsc{h}y\textsc{-gcn}} & 87.45 & 44.90 \\
         & \multicolumn{1}{l}{\textsc{h}y\textsc{-gat}} & 85.56 & 43.09 \\\cmidrule{2-4}
         &  \textbf{\textsc{\ourmodel{}}} &  \\
         & \multicolumn{1}{l}{\textsc{sage}} & \underline{93.32} & 58.34\\
         & \multicolumn{1}{l}{\textsc{gcn}} & \underline{93.61} & 58.65 \\
         & \multicolumn{1}{l}{\textsc{gat}} & \underline{93.65} & 58.55 \\
         & \multicolumn{1}{l}{\textsc{r-gcn}} & $\mathbf{94.69}$ &  \underline{61.71} \\
         & \multicolumn{1}{l}{\textsc{r-gat}} & $\mathbf{94.53}$ &  \underline{62.54} \\
         & \multicolumn{1}{l}{\textsc{h}y\textsc{-gcn}} & \underline{93.64} &  \underline{61.81} \\
         & \multicolumn{1}{l}{\textsc{h}y\textsc{-gat}} & \underline{92.97} &  \underline{61.91} \\\bottomrule
    \end{tabular}
    \caption{F1 scores (fake class) on \textit{GossipCop} and \textit{HealthStory}. ($^\dagger$) denotes results reported from \citet{defend} and ($^\ddagger$) from \citet{safe}. \textbf{Bold} figure denotes significantly better than other methods for that dataset. \underline{Underscore} figures denote significantly better scores than baselines but not significantly different from each other.}
    \label{tab: results}
\end{centering}
\end{table}

\vspace{1mm}
\noindent
\textbf{Results.} The mean F1 scores for all models are summarized in Table \ref{tab: results}. We note that the simple majority sharing baseline achieves an F1 of 77.19 on GossipCop while just 8.20 on HealthStory. This highlights the difference in the content sharing behavior of users between the two datasets and we explore this further in Section \ref{sec:content_share_behav}. We can also see this difference in the strength of social context information between the 2 datasets from the performance of the social baseline. Social baseline variants of all GNNs significantly ($p<0.05$ under \textit{paired t-test}) outperform all text-based methods in case of GossipCop but not in case of HealthStory. However, all the social baselines outperform the majority sharing baseline demonstrating the contribution of GNNs beyond capturing just the average sharing behavior of interacting users. Note that we observe similar trends in experiments with CNN as the text-encoder of the proposed framework.

Finally, in case of GossipCop, all the variants of the proposed \ourmodel{} framework significantly outperform all their social baseline counterparts as well as all the text-based models. The relational GNN variants significantly outperform all the methods, while the hyperbolic variants perform on par with the traditional GNNs. In case of HealthStory, we see that the traditional GNN variants significantly outperform their social baseline counterparts but not the best-performing text-based baseline (i.e., RoBERTa). However, the relational and hyperbolic GNNs significantly outperform all other methods. 

Overall, we see that the proposed relational GNNs outperform the traditional GNN models, indicating the importance of modelling the different relations between nodes of a heterogeneous graph separately. Hyperbolic GNNs are more expressive in embedding graphs that have a (deep) hierarchical structure. Due to the nature of the datasets and limitation of Twitter API (all retweets are mapped to the same source tweet, rather than forming a tree structure), the community graph is just 2 levels deep. Thus, hyperbolic GNNs perform similar to the traditional GNNs under our 2-layer setup. However, if more social information were available, resulting in a deeper graph, we expect Hy-GNNs to exhibit a superior performance. 

\begin{figure}[t!]
    \centering
        \hspace*{-0.2cm}\includegraphics[width=0.4\textwidth]{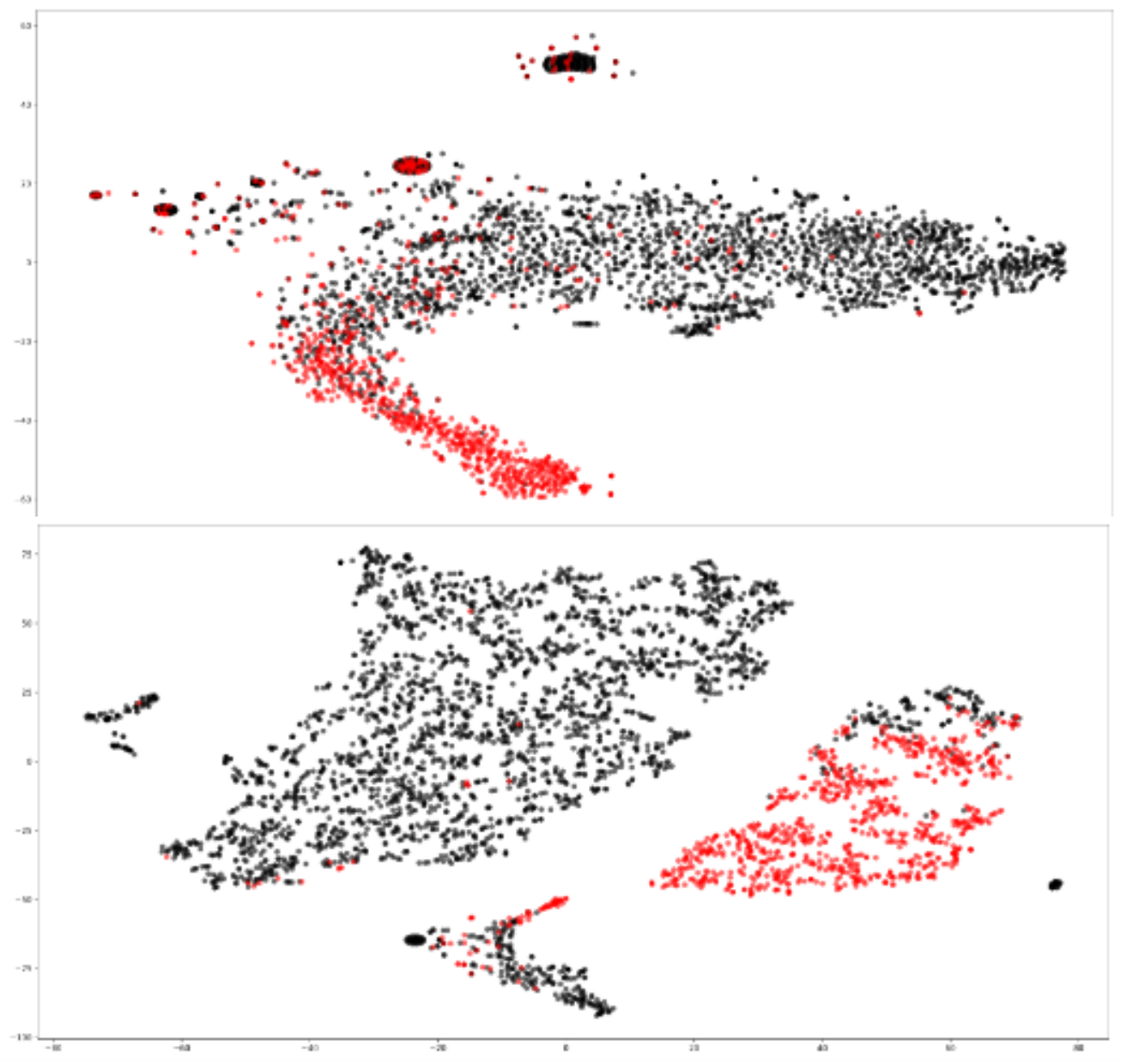}
    \caption{\textit{t}-SNE plots of test article embeddings produced by RoBERTa alone (top) and \ourmodel{} (R-GCN) (bottom). Fake articles are in red and real in black}
    \label{fig:tsne}
\end{figure}

In Figure \ref{fig:tsne}, we use \textit{t}-SNE \cite{tsne} to visualize the test articles representations generated by RoBERTa and \ourmodel{} (R-GCN). We see a much cleaner and compact segregation of fake and real classes when the community features are combined with the textual features of the articles compared to using textual features alone.



\section{Analysis}
\subsection{Effects of graph sparsity and frequent users}

Graph sparsity can affect the performance of GNNs as they rely on node connections to share information during training. Additionally, the presence of frequent users that share many articles of a particular class may introduce a bias in the model. In such cases, the network may learn to simply map a user to a class and use that as a shortcut for classification. To investigate the effects of these phenomena, we perform an ablation experiment on GossipCop by removing the more frequent/active users from the graph in a step-wise fashion. This makes the graph more sparse and discards many connections that the network could have learned to overfit on. Table \ref{tab: ablation} shows the performance of the GNN models when users sharing more than 10\%, 5\% and 1\% articles of \textit{each} class are removed from the graph. We see that the performance drops as users are removed successively; however, \ourmodel{} still outperforms all the text-based methods under the 10\% and 5\% setting, even without the presence of a possible bias introduced by frequent users. For the 1\% setting, only the hyperbolic GNNs outperform the baselines and this setting illustrates that under extremely sparse conditions (65\% of original density of an already sparse graph), the R-GNNs struggle to learn informative user representations. Overall, we see that Hy-GNNs are resilient to user biases (if any) and can perform well even on sparse graphs.

\begin{table}[t!]
\footnotesize
\begin{centering}
    \begin{tabular}{@{}lccc@{}}
        \toprule
        \textbf{Setting} & $\rho$ & \textbf{\ourmodel{}} & \textbf{F1} \\ \midrule
        \multirow{4}{*}{excl.\textgreater 10\%} & \multirow{4}{*}{0.78} &  \textsc{r-gcn} & 81.87 \\
        &  & \textsc{r-gat} & 82.27 \\
        &  & \textsc{h}y\textsc{-gcn} & 82.16 \\
        &  & \textsc{h}y\textsc{-gat} & 81.81 \\\midrule
        \multirow{4}{*}{excl.\textgreater 5\%} & \multirow{4}{*}{0.71} &  \textsc{r-gcn} & 77.16 \\
        &  & \textsc{r-gat} & 77.32 \\
        &  & \textsc{h}y\textsc{-gcn} & 77.13 \\
        &  & \textsc{h}y\textsc{-gat} & 77.01 \\\midrule
        \multirow{4}{*}{excl.\textgreater 1\%} & \multirow{4}{*}{0.65} & \textsc{r-gcn} & 65.89 \\
        &  & \textsc{r-gat} & 65.32 \\
        &  & \textsc{h}y\textsc{-gcn} & 71.99 \\
        &  & \textsc{h}y\textsc{-gat} & 72.05 \\\bottomrule  
    \end{tabular}
    \caption{Results of \ourmodel{} variants on varying subsets of user nodes on GossipCop. $\rho$ denotes relative graph density.}
    \label{tab: ablation}
\end{centering}
\end{table}

\subsection{Optimum support for effective learning}
\label{sec:optim_support}

GNNs learn node representations by aggregating information from their local neighborhoods. If the unsupervised nodes have very sparse connections (e.g., users that have shared just one (or very few) article(s)), then there is not enough support to learn their social-context features from. The effective neighborhood that a node uses is determined by the number of successive iterations of message passing steps, i.e., the number of layers in a GNN. Thus, in principle we can add more GNN layers to enable the sparse unsupervised nodes to gain access to back-propagating losses from distant supervised nodes too. However, simply stacking more layers leads to various known problems of training deep neural networks in general (vanishing gradients and overfitting due to large no. of parameters), as well as graph specific problems such as \textit{over-smoothing} and \textit{bottleneck phenomenon}.


\textit{Over-smoothing} is the phenomenon where node features tend to converge to the same vector and become nearly indistinguishable as the result of applying multiple GNN layers \cite{ovr_smooth_1, ovr_smooth_3}. Moreover, in social network graphs, predictions typically rely only on short-range information from the local neighbourhood of a node and do not improve by adding distant information. In our community graph, modeling information from 2 hops away is sufficient to aggregate useful community-wide information at each node and can be achieved with 2-layer GNNs. Thus, learning node representations for sparsely connected nodes from these shallow GNNs is challenging. 

\textit{Bottleneck} is the phenomenon of ``over-squashing” of information from exponentially many neighbours into small fixed-size vectors \cite{gnn_bottleneck}. Since each article is shared by many users and each user is connected to many other users, the network can suffer from bottleneck which affects learning. In a 2-layer GNN setup, the effective aggregation neighborhoods of each article node exponentially increases, as it aggregates information from all the nodes that are within 2-hops away from it.

\begin{figure}[t!]
    \centering
        \includegraphics[width=0.35\textwidth]{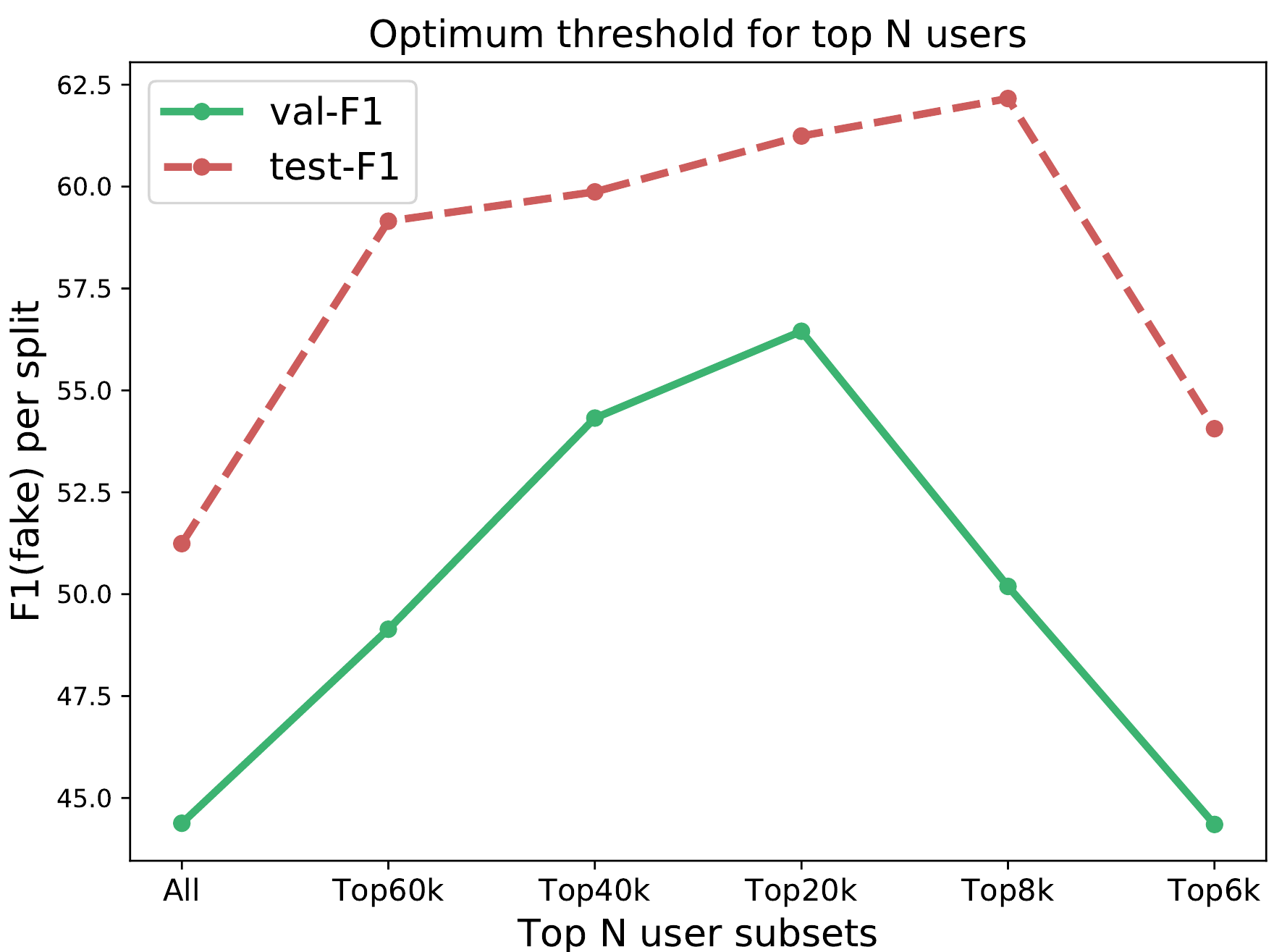}
    \caption[Optimum threshold for effective learning]{Validation and test set performance of the \ourmodel{} (GCN) framework over varying subsets of most active users on HealthStory.}
    \label{fig:optim_thresh}
\end{figure}

Due to these observations during our initial experiments, we choose to use just the ``top N most active users".
We define ``active users" as those that have shared more articles, i.e., have sufficient support to learn from, and hence can help us capture their content-sharing behavior better. In Figure \ref{fig:optim_thresh} we show the validation and test performance over varying subsets of active users in HealthStory. We see that as we successively drop the least active users, the validation and test scores show a positive trend. This illustrates the effects of bottleneck on the network. However, the scores drop after a certain threshold of users. This threshold is the optimum number of users required to learn effectively using the GNNs -- adding more users leads to bottleneck and removing users leads to underfitting due to the lack of sufficient support to learn from. We see that validation and test scores are correlated in this behavior and we tune our optimal threshold of users for effective learning based on the validation set of \ourmodel{} (GCN) and use the same subset for all the other GNN encoders for fair comparison.
The best validation score was achieved at the \textit{top20K} subset of most active users while the test scores peaked for the \textit{top8K} setting. Thus, we run all our experiments with the top 20K active users for HealthStory, and similarly top30K for GossipCop.

\subsection{Effect of article sharing patterns}
\label{sec:content_share_behav}

\begin{figure}[t!]
    \centering
        \includegraphics[width=0.5\textwidth]{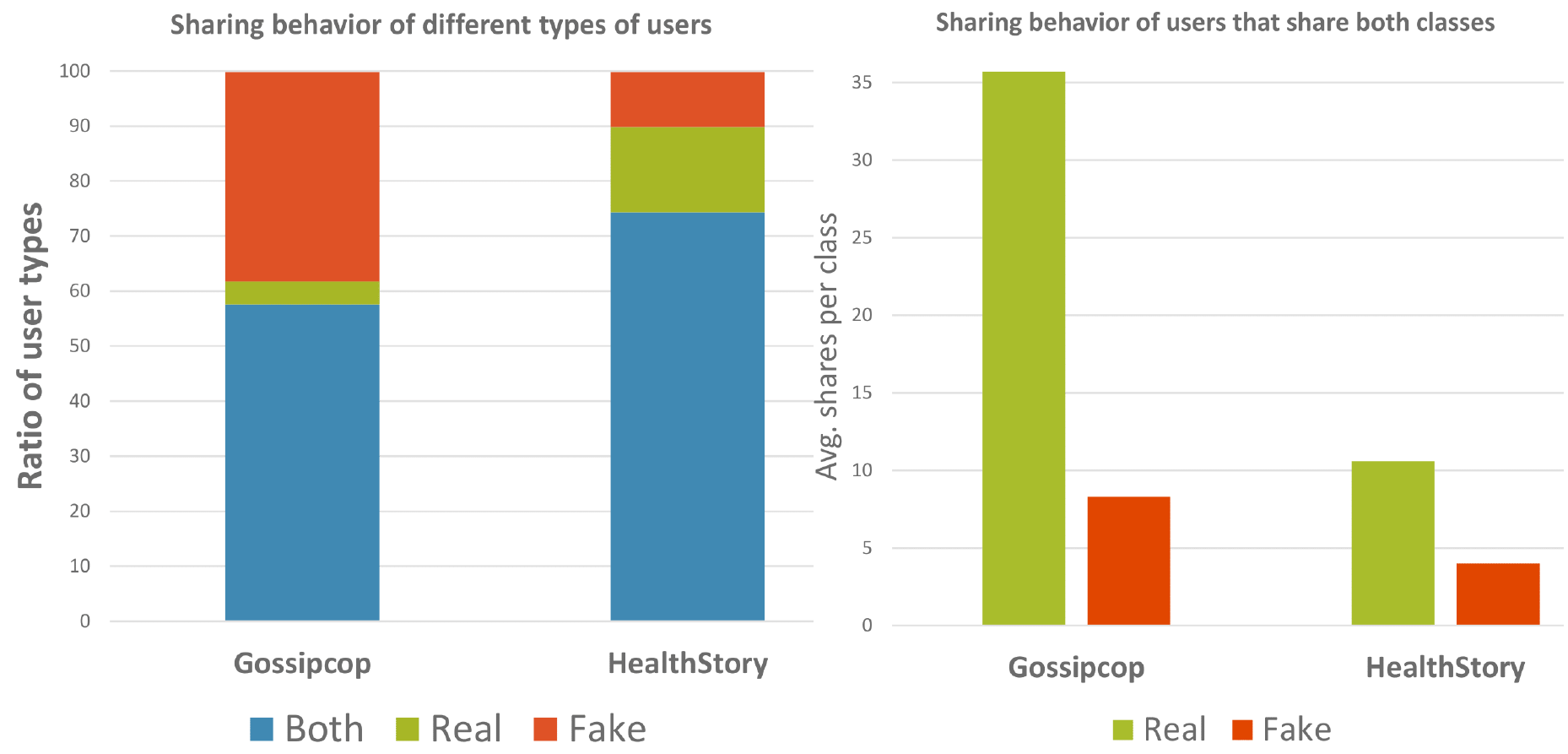}
    \caption{Article sharing behavior of 3 kinds of users (left) and; Average of real and fake article shares of type (c) users (right).}
    \label{fig:share_behav}
\end{figure}

As discussed earlier, results in Table \ref{tab: results} show that there is a difference in the article sharing behavior of users between the two datasets. To understand user characteristics better, we visualize the article sharing behavior of users for both the datasets in Figure \ref{fig:share_behav}. We visualize the composition of 3 types of users in the datasets: (a) users that share only real articles, (b) only fake articles and (c) those that share articles from both classes. We see that the majority of the users are type (c) in both datasets (57.18\% for GossipCop and 74.15\% for HealthStory). However, 38\% of the users are type (b) in GossipCop while just 9.96\% in HealthStory. Furthermore, we visualize the average of real and fake articles shared by type (c) users on the right in Figure \ref{fig:share_behav}. From these observations, we note that the GNNs are better positioned to learn user representations to detect fake articles in case of GossipCop, since: (1) The community graph has enough support of type (b) users (38\%). This aids the GNNs to learn rich community-level features of users that aid in detecting fake articles; (2) Even of the 57\% type (c) users, they are much more likely to share articles of a single class (here, real). This again helps the network to learn distinct features for these users and assign them to a specific community.


However, in case of HealthStory, the GNNs struggle to learn equally rich user representations to detect fake articles since: (1) The community graph has only around 10\% of type (b) users. This limits the GNNs from learning expressive community level features for users that are more likely to share fake articles and thereby are not able to use them for accurate prediction. (2) A vast majority of users (74\%) share articles of both classes. To add to that, these bulk of users are considerably less likely to share articles of one class predominantly. This again restricts the GNNs from learning informative representations of these users, as it struggles to assign them to any specific community due to mixed signals.


\section{Conclusion}
We presented a graph-based approach to fake news detection which leverages information-spreading behaviour of social media users. Our results demonstrate that incorporating community-based modeling leads to substantially improved performance in this task as compared to purely text-based models. The proposed relational GNNs for user/community modeling outperformed the traditional GNNs indicating the importance of explicitly modeling the relations in a heterogeneous graph. Meanwhile, the proposed hyperbolic GNNs performed on par with other GNNs and we leave their application for user/community modeling to truly hierarchical social network datasets as future work. 
In the future, it would be interesting to apply these techniques to other tasks, such as rumour detection and modeling changes in public beliefs.


\newpage
\bibliographystyle{acl_natbib}
\bibliography{emnlp2020}

\begin{thebibliography}{53}
\expandafter\ifx\csname natexlab\endcsname\relax\def\natexlab#1{#1}\fi

\bibitem[{Alon and Yahav(2020)}]{gnn_bottleneck}
Uri Alon and Eran Yahav. 2020.
\newblock On the bottleneck of graph neural networks and its practical
  implications.
\newblock \emph{arXiv preprint arXiv:2006.05205}.

\bibitem[{Balmas(2014)}]{fakenews_effects1}
Meital Balmas. 2014.
\newblock When fake news becomes real: Combined exposure to multiple news
  sources and political attitudes of inefficacy, alienation, and cynicism.
\newblock \emph{Communication research}, 41(3):430--454.

\bibitem[{B{\'e}cigneul and Ganea(2018)}]{radam}
Gary B{\'e}cigneul and Octavian-Eugen Ganea. 2018.
\newblock Riemannian adaptive optimization methods.
\newblock \emph{arXiv preprint arXiv:1810.00760}.

\bibitem[{Castillo et~al.(2011)Castillo, Mendoza, and
  Poblete}]{tfidf_style_based}
Carlos Castillo, Marcelo Mendoza, and Barbara Poblete. 2011.
\newblock Information credibility on twitter.
\newblock In \emph{Proceedings of the 20th international conference on World
  wide web}, pages 675--684.

\bibitem[{Chami et~al.(2019)Chami, Ying, Ré, and Leskovec}]{hygcn}
Ines Chami, Rex Ying, Christopher Ré, and Jure Leskovec. 2019.
\newblock \href {http://arxiv.org/abs/1910.12933} {Hyperbolic graph
  convolutional neural networks}.

\bibitem[{Chen et~al.(2013)Chen, Fang, Hu, and Mahoney}]{eucl_distort2}
Wei Chen, Wenjie Fang, Guangda Hu, and Michael~W Mahoney. 2013.
\newblock On the hyperbolicity of small-world and treelike random graphs.
\newblock \emph{Internet Mathematics}, 9(4):434--491.

\bibitem[{Chiang et~al.(2019)Chiang, Liu, Si, Li, Bengio, and
  Hsieh}]{cluster_gcn}
Wei-Lin Chiang, Xuanqing Liu, Si~Si, Yang Li, Samy Bengio, and Cho-Jui Hsieh.
  2019.
\newblock Cluster-gcn: An efficient algorithm for training deep and large graph
  convolutional networks.
\newblock In \emph{Proceedings of the 25th ACM SIGKDD International Conference
  on Knowledge Discovery \& Data Mining}, pages 257--266.

\bibitem[{Ciampaglia et~al.(2015)Ciampaglia, Shiralkar, Rocha, Bollen, Menczer,
  and Flammini}]{spo_1}
Giovanni~Luca Ciampaglia, Prashant Shiralkar, Luis~M Rocha, Johan Bollen,
  Filippo Menczer, and Alessandro Flammini. 2015.
\newblock Computational fact checking from knowledge networks.
\newblock \emph{PloS one}, 10(6).

\bibitem[{Dai et~al.(2020)Dai, Sun, and Wang}]{fakehealth}
Enyan Dai, Yiwei Sun, and Suhang Wang. 2020.
\newblock Ginger cannot cure cancer: Battling fake health news with a
  comprehensive data repository.
\newblock In \emph{Proceedings of the International AAAI Conference on Web and
  Social Media}, volume~14, pages 853--862.

\bibitem[{Del~Tredici et~al.(2019)Del~Tredici, Marcheggiani, Schulte~im Walde,
  and Fern{\'a}ndez}]{gat_stance}
Marco Del~Tredici, Diego Marcheggiani, Sabine Schulte~im Walde, and Raquel
  Fern{\'a}ndez. 2019.
\newblock \href {https://www.aclweb.org/anthology/D19-1477} {You shall know a
  user by the company it keeps: Dynamic representations for social media users
  in {NLP}}.
\newblock In \emph{Proceedings of the 2019 Conference on Empirical Methods in
  Natural Language Processing and the 9th International Joint Conference on
  Natural Language Processing (EMNLP-IJCNLP)}, pages 4707--4717. ACL.

\bibitem[{Fr{\'e}chet(1948)}]{frechet_mean}
Maurice Fr{\'e}chet. 1948.
\newblock Les {\'e}l{\'e}ments al{\'e}atoires de nature quelconque dans un
  espace distanci{\'e}.
\newblock In \emph{Annales de l'institut Henri Poincar{\'e}}, volume~10, pages
  215--310.

\bibitem[{Ganea et~al.(2018)Ganea, B{\'e}cigneul, and Hofmann}]{hnn2}
Octavian Ganea, Gary B{\'e}cigneul, and Thomas Hofmann. 2018.
\newblock Hyperbolic neural networks.
\newblock In \emph{Advances in neural information processing systems}, pages
  5345--5355.

\bibitem[{Hamilton et~al.(2017)Hamilton, Ying, and Leskovec}]{graphsage}
Will Hamilton, Zhitao Ying, and Jure Leskovec. 2017.
\newblock Inductive representation learning on large graphs.
\newblock In \emph{Advances in neural information processing systems}, pages
  1024--1034.

\bibitem[{Han et~al.(2020)Han, Karunasekera, and
  Leckie}]{gnn_continual_learning}
Yi~Han, Shanika Karunasekera, and Christopher Leckie. 2020.
\newblock \href {http://arxiv.org/abs/2007.03316} {Graph neural networks with
  continual learning for fake news detection from social media}.

\bibitem[{Karypis and Kumar(1998)}]{metis}
George Karypis and Vipin Kumar. 1998.
\newblock A fast and high quality multilevel scheme for partitioning irregular
  graphs.
\newblock \emph{SIAM Journal on scientific Computing}, 20(1):359--392.

\bibitem[{Kashima et~al.(2003)Kashima, Tsuda, and
  Inokuchi}]{graph_kernels_orig2}
Hisashi Kashima, Koji Tsuda, and Akihiro Inokuchi. 2003.
\newblock Marginalized kernels between labeled graphs.
\newblock In \emph{Proceedings of the 20th ICML (ICML-03)}, pages 321--328.

\bibitem[{Khan et~al.(2019)Khan, Khondaker, Islam, Iqbal, and Afroz}]{conv_han}
Junaed~Younus Khan, Md~Khondaker, Tawkat Islam, Anindya Iqbal, and Sadia Afroz.
  2019.
\newblock A benchmark study on machine learning methods for fake news
  detection.
\newblock \emph{arXiv preprint arXiv:1905.04749}.

\bibitem[{Kim(2014)}]{kim-cnn}
Yoon Kim. 2014.
\newblock \href {https://www.aclweb.org/anthology/D14-1181} {Convolutional
  neural networks for sentence classification}.
\newblock In \emph{Proceedings of the 2014 Conference on Empirical Methods in
  Natural Language Processing ({EMNLP})}, pages 1746--1751. ACL.

\bibitem[{Kipf and Welling(2016)}]{gcn}
Thomas~N Kipf and Max Welling. 2016.
\newblock Semi-supervised classification with graph convolutional networks.
\newblock \emph{arXiv preprint arXiv:1609.02907}.

\bibitem[{Li and Goldwasser(2019)}]{pol_persp_gcn}
Chang Li and Dan Goldwasser. 2019.
\newblock \href {https://www.aclweb.org/anthology/P19-1247} {Encoding social
  information with graph convolutional networks for{P}olitical perspective
  detection in news media}.
\newblock In \emph{Proceedings of the 57th Annual Meeting of the ACL}, pages
  2594--2604. ACL.

\bibitem[{Liu et~al.(2019{\natexlab{a}})Liu, Nickel, and Kiela}]{hnn}
Qi~Liu, Maximilian Nickel, and Douwe Kiela. 2019{\natexlab{a}}.
\newblock \href
  {http://papers.nips.cc/paper/9033-hyperbolic-graph-neural-networks.pdf}
  {Hyperbolic graph neural networks}.
\newblock In H.~Wallach, H.~Larochelle, A.~Beygelzimer, F.~d\' Alch\'{e}-Buc,
  E.~Fox, and R.~Garnett, editors, \emph{Advances in Neural Information
  Processing Systems 32}, pages 8230--8241. Curran Associates, Inc.

\bibitem[{Liu et~al.(2019{\natexlab{b}})Liu, Ott, Goyal, Du, Joshi, Chen, Levy,
  Lewis, Zettlemoyer, and Stoyanov}]{roberta}
Yinhan Liu, Myle Ott, Naman Goyal, Jingfei Du, Mandar Joshi, Danqi Chen, Omer
  Levy, Mike Lewis, Luke Zettlemoyer, and Veselin Stoyanov. 2019{\natexlab{b}}.
\newblock Roberta: A robustly optimized bert pretraining approach.
\newblock \emph{arXiv preprint arXiv:1907.11692}.

\bibitem[{Loshchilov and Hutter(2017)}]{adamw}
Ilya Loshchilov and Frank Hutter. 2017.
\newblock \href {http://arxiv.org/abs/1711.05101} {Decoupled weight decay
  regularization}.

\bibitem[{Ma et~al.(2016)Ma, Gao, Mitra, Kwon, Jansen, Wong, and
  Cha}]{simple_rnn_fakenews}
Jing Ma, Wei Gao, Prasenjit Mitra, Sejeong Kwon, Bernard~J. Jansen, Kam-Fai
  Wong, and Meeyoung Cha. 2016.
\newblock Detecting rumors from microblogs with recurrent neural networks.
\newblock In \emph{Proceedings of the Twenty-Fifth International Joint
  Conference on Artificial Intelligence}, IJCAI’16, page 3818–3824. AAAI
  Press.

\bibitem[{Ma et~al.(2018)Ma, Gao, and Wong}]{recursive_nn}
Jing Ma, Wei Gao, and Kam-Fai Wong. 2018.
\newblock \href {https://www.aclweb.org/anthology/P18-1184} {Rumor detection on
  twitter with tree-structured recursive neural networks}.
\newblock In \emph{Proceedings of the 56th Annual Meeting of the ACL (Volume 1:
  Long Papers)}, pages 1980--1989. ACL.

\bibitem[{Maaten and Hinton(2008)}]{tsne}
Laurens van~der Maaten and Geoffrey Hinton. 2008.
\newblock Visualizing data using t-sne.
\newblock \emph{Journal of machine learning research}, 9(Nov):2579--2605.

\bibitem[{Mishra et~al.(2019)Mishra, Del~Tredici, Yannakoudakis, and
  Shutova}]{abuse_detection}
Pushkar Mishra, Marco Del~Tredici, Helen Yannakoudakis, and Ekaterina Shutova.
  2019.
\newblock \href {https://www.aclweb.org/anthology/N19-1221} {{A}busive
  {L}anguage {D}etection with {G}raph {C}onvolutional {N}etworks}.
\newblock In \emph{Proceedings of the 2019 Conference of the North {A}merican
  Chapter of the ACL: Human Language Technologies, Volume 1 (Long and Short
  Papers)}, pages 2145--2150. ACL.

\bibitem[{Mishra et~al.(2020)Mishra, Piktus, Goossen, and
  Silvestri}]{mishra2020node}
Pushkar Mishra, Aleksandra Piktus, Gerard Goossen, and Fabrizio Silvestri.
  2020.
\newblock \href {https://arxiv.org/pdf/2001.07524.pdf} {Node masking: Making
  graph neural networks generalize and scale better}.
\newblock \emph{ArXiv}, abs/2001.07524.

\bibitem[{Nguyen et~al.(2020)Nguyen, Sugiyama, Nakov, and Kan}]{fang}
Van-Hoang Nguyen, Kazunari Sugiyama, Preslav Nakov, and Min-Yen Kan. 2020.
\newblock \href {http://arxiv.org/abs/2008.07939} {Fang: Leveraging social
  context for fake news detection using graph representation}.

\bibitem[{Norton and Greenwald(2016)}]{fakenews_effects2}
Ben Norton and Glenn Greenwald. 2016.
\newblock \href
  {{https://theintercept.com/2016/11/26/washington-post-disgracefully-promotes-a-mccarthyite-blacklist-from-a-new-hidden-and-very-shady-group/}}
  {{Washington Post disgracefully promotes a McCarthyite Blacklist from a
  hidden, new and very shady group}}.

\bibitem[{NT and Maehara(2019)}]{ovr_smooth_3}
Hoang NT and Takanori Maehara. 2019.
\newblock Revisiting graph neural networks: All we have is low-pass filters.
\newblock \emph{arXiv preprint arXiv:1905.09550}.

\bibitem[{Okano et~al.(2020)Okano, Liu, Ji, and Ruiz}]{han_fakenews}
Emerson~Yoshiaki Okano, Zebin Liu, Donghong Ji, and Evandro Eduardo~Seron Ruiz.
  2020.
\newblock Fake news detection on fake.br using hierarchical attention networks.
\newblock In \emph{Computational Processing of the Portuguese Language}, pages
  143--152. Springer International Publishing.

\bibitem[{Oono and Suzuki(2019)}]{ovr_smooth_1}
Kenta Oono and Taiji Suzuki. 2019.
\newblock Graph neural networks exponentially lose expressive power for node
  classification.
\newblock \emph{arXiv preprint arXiv:1905.10947}.

\bibitem[{P{\'e}rez-Rosas et~al.(2017)P{\'e}rez-Rosas, Kleinberg, Lefevre, and
  Mihalcea}]{style_based_auto_detect}
Ver{\'o}nica P{\'e}rez-Rosas, Bennett Kleinberg, Alexandra Lefevre, and Rada
  Mihalcea. 2017.
\newblock Automatic detection of fake news.
\newblock \emph{arXiv preprint arXiv:1708.07104}.

\bibitem[{Peters et~al.(2018)Peters, Neumann, Iyyer, Gardner, Clark, Lee, and
  Zettlemoyer}]{elmo}
Matthew Peters, Mark Neumann, Mohit Iyyer, Matt Gardner, Christopher Clark,
  Kenton Lee, and Luke Zettlemoyer. 2018.
\newblock \href {https://www.aclweb.org/anthology/N18-1202} {Deep
  contextualized word representations}.
\newblock In \emph{Proceedings of the 2018 Conference of the North {A}merican
  Chapter of the ACL: Human Language Technologies, Volume 1 (Long Papers)},
  pages 2227--2237. ACL.

\bibitem[{Popat(2017)}]{lang_style}
Kashyap Popat. 2017.
\newblock Assessing the credibility of claims on the web.
\newblock In \emph{Proceedings of the 26th International Conference on World
  Wide Web Companion}, pages 735--739.

\bibitem[{Potthast et~al.(2017)Potthast, Kiesel, Reinartz, Bevendorff, and
  Stein}]{style_based_stylometric}
Martin Potthast, Johannes Kiesel, Kevin Reinartz, Janek Bevendorff, and Benno
  Stein. 2017.
\newblock A stylometric inquiry into hyperpartisan and fake news.
\newblock \emph{arXiv preprint arXiv:1702.05638}.

\bibitem[{Ravasz and Barab{\'a}si(2003)}]{eucl_distort1}
Erzs{\'e}bet Ravasz and Albert-L{\'a}szl{\'o} Barab{\'a}si. 2003.
\newblock Hierarchical organization in complex networks.
\newblock \emph{Physical review E}, 67(2):026112.

\bibitem[{Schlichtkrull et~al.(2018)Schlichtkrull, Kipf, Bloem, Van Den~Berg,
  Titov, and Welling}]{rgcn}
Michael Schlichtkrull, Thomas~N Kipf, Peter Bloem, Rianne Van Den~Berg, Ivan
  Titov, and Max Welling. 2018.
\newblock Modeling relational data with graph convolutional networks.
\newblock In \emph{European Semantic Web Conference}, pages 593--607. Springer.

\bibitem[{Shi and Weninger(2016)}]{spo_2}
Baoxu Shi and Tim Weninger. 2016.
\newblock Discriminative predicate path mining for fact checking in knowledge
  graphs.
\newblock \emph{Knowledge-based systems}, 104:123--133.

\bibitem[{Shu et~al.(2019{\natexlab{a}})Shu, Cui, Wang, Lee, and Liu}]{defend}
Kai Shu, Limeng Cui, Suhang Wang, Dongwon Lee, and Huan Liu.
  2019{\natexlab{a}}.
\newblock \href {https://doi.org/10.1145/3292500.3330935} {Defend: Explainable
  fake news detection}.
\newblock In \emph{Proceedings of the 25th ACM SIGKDD International Conference
  on Knowledge Discovery \& Data Mining}, KDD ’19, page 395–405.
  Association for Computing Machinery.

\bibitem[{Shu et~al.(2018)Shu, Mahudeswaran, Wang, Lee, and Liu}]{fakenewsnet}
Kai Shu, Deepak Mahudeswaran, Suhang Wang, Dongwon Lee, and Huan Liu. 2018.
\newblock Fakenewsnet: A data repository with news content, social context and
  dynamic information for studying fake news on social media.
\newblock \emph{arXiv preprint arXiv:1809.01286}.

\bibitem[{Shu et~al.(2019{\natexlab{b}})Shu, Wang, and Liu}]{social_cntxtt_1}
Kai Shu, Suhang Wang, and Huan Liu. 2019{\natexlab{b}}.
\newblock Beyond news contents: The role of social context for fake news
  detection.
\newblock In \emph{Proceedings of the Twelfth ACM International Conference on
  Web Search and Data Mining}, pages 312--320.

\bibitem[{Shu et~al.(2020)Shu, Zheng, Li, Mukherjee, Awadallah, Ruston, and
  Liu}]{social_cntxt_2}
Kai Shu, Guoqing Zheng, Yichuan Li, Subhabrata Mukherjee, Ahmed~Hassan
  Awadallah, Scott Ruston, and Huan Liu. 2020.
\newblock Leveraging multi-source weak social supervision for early detection
  of fake news.
\newblock \emph{arXiv preprint arXiv:2004.01732}.

\bibitem[{Shu et~al.(2019{\natexlab{c}})Shu, Zhou, Wang, Zafarani, and
  Liu}]{multi_modal_1}
Kai Shu, Xinyi Zhou, Suhang Wang, Reza Zafarani, and Huan Liu.
  2019{\natexlab{c}}.
\newblock The role of user profiles for fake news detection.
\newblock In \emph{Proceedings of the 2019 IEEE/ACM International Conference on
  Advances in Social Networks Analysis and Mining}, pages 436--439.

\bibitem[{Srivastava et~al.(2014)Srivastava, Hinton, Krizhevsky, Sutskever, and
  Salakhutdinov}]{dropout}
Nitish Srivastava, Geoffrey Hinton, Alex Krizhevsky, Ilya Sutskever, and Ruslan
  Salakhutdinov. 2014.
\newblock Dropout: a simple way to prevent neural networks from overfitting.
\newblock \emph{The journal of machine learning research}, 15(1):1929--1958.

\bibitem[{Veli{\v{c}}kovi{\'c} et~al.(2017)Veli{\v{c}}kovi{\'c}, Cucurull,
  Casanova, Romero, Lio, and Bengio}]{gat}
Petar Veli{\v{c}}kovi{\'c}, Guillem Cucurull, Arantxa Casanova, Adriana Romero,
  Pietro Lio, and Yoshua Bengio. 2017.
\newblock Graph attention networks.
\newblock \emph{arXiv preprint arXiv:1710.10903}.

\bibitem[{Wang et~al.(2018)Wang, Ma, Jin, Yuan, Xun, Jha, Su, and Gao}]{eann}
Yaqing Wang, Fenglong Ma, Zhiwei Jin, Ye~Yuan, Guangxu Xun, Kishlay Jha, Lu~Su,
  and Jing Gao. 2018.
\newblock Eann: Event adversarial neural networks for multi-modal fake news
  detection.
\newblock In \emph{Proceedings of the 24th acm sigkdd international conference
  on knowledge discovery \& data mining}, pages 849--857. ACM.

\bibitem[{Wolf et~al.(2019)Wolf, Debut, Sanh, Chaumond, Delangue, Moi, Cistac,
  Rault, Louf, Funtowicz, and Brew}]{hugging_face}
Thomas Wolf, Lysandre Debut, Victor Sanh, Julien Chaumond, Clement Delangue,
  Anthony Moi, Pierric Cistac, Tim Rault, R'emi Louf, Morgan Funtowicz, and
  Jamie Brew. 2019.
\newblock Huggingface's transformers: State-of-the-art natural language
  processing.
\newblock \emph{ArXiv}, abs/1910.03771.

\bibitem[{Wu et~al.(2015)Wu, Yang, and Zhu}]{cascade_similarity}
Ke~Wu, Song Yang, and Kenny~Q Zhu. 2015.
\newblock False rumors detection on sina weibo by propagation structures.
\newblock In \emph{2015 IEEE 31st international conference on data
  engineering}, pages 651--662. IEEE.

\bibitem[{Zarocostas(2020)}]{Infodemic}
John Zarocostas. 2020.
\newblock {How to fight an Infodemic}.
\newblock \emph{The Lancet}, 395(10255):676.

\bibitem[{Zhou et~al.(2020)Zhou, Wu, and Zafarani}]{safe}
Xinyi Zhou, Jindi Wu, and Reza Zafarani. 2020.
\newblock Safe: Similarity-aware multi-modal fake news detection.
\newblock \emph{arXiv preprint arXiv:2003.04981}.

\bibitem[{Zhou and Zafarani(2018)}]{fakenews_survey}
Xinyi Zhou and Reza Zafarani. 2018.
\newblock Fake news: A survey of research, detection methods, and
  opportunities.
\newblock \emph{arXiv preprint arXiv:1812.00315}.

\end{thebibliography}
\clearpage
\newpage

\appendix
\section{Appendix}

\subsection{Text preprocessing}
We clean the raw text of the crawled articles of the \textit{GossipCop} dataset before using them for training. More specifically, we replace any URLs and hashtags in the text with the tokens \texttt{[url]} and \texttt{[hashtag]} respectively. We also replace new line characters with a blank space and make sure that class distributions across the train-val-test splits are the same. 

\subsection{Hyper-parameters}
All our code is in PyTorch and we use the HuggingFace library \cite{hugging_face} to train the transformer models. We grid-search over the following values of the parameters for the respective models and choose the best setting based on best F1 score on test set:

\begin{enumerate}
    \item \textbf{CNN:} learning rate = [5e-3, 1e-3, 5e-4, 1e-4], dropout = [0.1, 0.2, 0.3, 0.4, 0.5, 0.6], weight decay = [1e-3,2e-3]
    
    \item \textbf{Transformers:} learning rate = [5e-3, 1e-3, 5e-4, 1e-4], weight decay = [1e-3, 1e-2, 1e-1, 5e-1], hidden dropout = [0.1, 0.2, 0.3, 0.4, 0.5], attention dropout = [0.1, 0.2, 0.3, 0.4, 0.5]
    
    \item \textbf{GNNs:} learning rate = [5e-3, 1e-3, 5e-4, 1e-4], weight decay = [1e-3, 2e-3], hidden dropout = [0.1, 0.2, 0.3, 0.4, 0.5], node mask = [0.1, 0.2, 0.3, 0.4, 0.5], hidden dimension = [128, 256, 512]
\end{enumerate}

The set of best hyper-parameters for all models are reported in Table \ref{tab: hyper-param}.


\begin{table*}
\begin{centering}
    \footnotesize
    \begin{tabular}{lccccccc|cc}
    \toprule
         & \multicolumn{7}{c|}{\textbf{Graph}}  & \multicolumn{2}{c}{\textbf{Text}}\\\cmidrule{2-10}
          & \textbf{\textsc{gcn}} & \textbf{\textsc{gat}} & \textbf{\textsc{sage}} & \textbf{\textsc{r-gcn}} & \textbf{\textsc{r-gat}} & \textbf{\textsc{h}y\textsc{-gcn}} & \textbf{\textsc{h}y\textsc{-gat}} & \textbf{\textsc{cnn}} & \textbf{\textsc{r}o\textsc{bert}a} \\\cmidrule{2-10}
         Learning rate & $5 \cdot10^{-4}$ & $5 \cdot10^{-4}$ & $1 \cdot10^{-4}$ & $1 \cdot10^{-3}$ & $1 \cdot10^{-4}$ & $5 \cdot10^{-3}$ & $5 \cdot10^{-3}$ & $5 \cdot10^{-4}$ & $5 \cdot10^{-4}$ \\
         Weight Decay & $1 \cdot10^{-3}$ & $1 \cdot10^{-3}$ & $2 \cdot10^{-3}$ & $2 \cdot10^{-3}$ & $1 \cdot10^{-3}$ & $1 \cdot10^{-3}$ & $1 \cdot10^{-3}$ & $1 \cdot10^{-3}$ & $5 \cdot10^{-1}$ \\
         Attention dropout & NA & 0.1 & NA & NA & 0.1 & NA & NA & NA & $0.4$ \\
         Hidden dropout & 0.1 & 0.4 & 0.2 & 0.4 & 0.2 & 0.1 & 0.1 & $0.5$ & $0.1$ \\
         Node masking prob. & 0.1 & 0.1 & 0.1 & 0.1 & 0.1 & 0.1 & 0.1 & NA & NA \\
         Hidden dimension & 256 & 512 & 128 & 512 & 512 & 256 & 512 & 384 & 1024 \\\midrule
         Learning rate & $5 \cdot10^{-3}$ & $5 \cdot10^{-4}$ & $1 \cdot10^{-4}$ & $5 \cdot10^{-4}$ & $1 \cdot10^{-3}$ & $5 \cdot10^{-3}$ & $5 \cdot10^{-3}$ & $5 \cdot10^{-4}$ & $5 \cdot10^{-4}$ \\
         Weight Decay & $1 \cdot10^{-3}$ & $2 \cdot10^{-3}$ & $2 \cdot10^{-3}$ & $2 \cdot10^{-3}$ & $1 \cdot10^{-3}$ & $1 \cdot10^{-3}$ & $1 \cdot10^{-3}$ & $1 \cdot10^{-3}$ & $5 \cdot10^{-1}$ \\
         Attention dropout & NA & NA & NA & NA & NA & NA & NA & NA & $0.4$ \\
         Hidden dropout & 0.4 & 0.2 & 0.2 & 0.2 & 0.2 & 0.1 & 0.1 & $0.5$ & $0.1$ \\
         Node masking prob. & 0.1 & 0.2 & 0.1 & 0.2 & 0.1 & 0.3 & 0.3 & NA & NA \\
         Hidden dimension & 512 & 512 & 128 & 512 & 512 & 512 & 256 & 384 & 1024 \\\bottomrule
    \end{tabular}
    \caption{Best Hyper-parameters for all the models on GossipCop (top) and HealthStory (bottom).}
    \label{tab: hyper-param}
\end{centering}
\end{table*}


\subsection{Hardware and Run Times}
We use NVIDIA Titanrtx 2080Ti for training multiple-GPU models and 1080ti for single GPU ones. In Table \ref{tab: runtime} we report the run times (per epoch) for each model.

\begin{table*}[t!]
\begin{centering}
    \footnotesize
    \begin{tabular}{c|cc|cc}
    \toprule
        & \multicolumn{2}{|c|}{\textbf{GossipCop}} & \multicolumn{2}{c}{\textbf{HealthStory}} \\ \midrule
          \textbf{Method} & \textbf{No. of GPUs} & \textbf{Run time (per epoch)} & \textbf{No. of GPUs} & \textbf{Run time (per epoch)} \\\midrule
         \textsc{cnn} & 4  & 15 mins & 4  & 1.25 mins\\
         \textsc{r}o\textsc{bert}a & 4 & 6 mins & 4 & 3 mins \\\midrule
         \textsc{sage} & 1 & 8.77 secs & 1 & 1.49 secs \\
         \textsc{gcn} & 1 & 6.06 secs & 1 & 1.91  secs\\
         \textsc{gat} & 1 & 6.76 secs & 1 & 1.96 secs \\
         \textsc{rgcn} & 1 & 6.92 secs & 1 & 1.40 secs \\
         \textsc{rgat} & 1 & 7.88 secs & 1 & 2.16 secs \\
         \textsc{h}y\textsc{-gcn} & 1 & 10.39 secs & 1 & 1.64 secs \\
         \textsc{h}y\textsc{-gat} & 1 & 16.50 secs & 1 & 2.97 secs \\\bottomrule
    \end{tabular}
    \caption{Per epoch run times of all the models}
    \label{tab: runtime}
\end{centering}
\end{table*}

\subsection{Evaluation Metric}

We use F1 score (of the target class, ie, fake class) to report all our performance. F1 is defined as : 
\begin{equation*}
    F1 = 2 \times \frac{Precision \times Recall}{Precision + Recall}
\end{equation*}

where, Precision and Recall are defined as:

\begin{equation*}
    Precision = \frac{True \ Positive}{True \ Positive + False \ Positive}
\end{equation*}

\begin{equation*}
    Recall = \frac{True \ Positive}{True \ Positive + False \ Negative}
\end{equation*}

\subsection{Training Details}

\begin{enumerate}
    \item To leverage effective batching of graph data during training, we cluster the Graph into 300 dense sub-graphs using the METIS \cite{metis} graph clustering algorithm. We then train all the GNN networks with a batch-size of 16, ie, 16 of these sub-graphs are sampled at each pass as detailed in \cite{cluster_gcn}. This vastly reduces the time, memory and computation complexity of large sparse graphs.
    
    \item Additionally, for GCN we adopt "diagonal enhancement" by adding identity to the original adjacency matrix $A$ \cite{cluster_gcn} and perform the normalization as:$\tilde{A} = (D+I)^{-1}(A+I)$.
    
    \item For SAGE we use "mean" aggregation and normalize the output features as $\frac{\mathbf{x}_{i}^{\prime}}{\left\|\mathbf{x}_{i}^{\prime}\right\|_{2}}$ where, $x^{\prime}_i$ is $\mathbf{x}_{i}^{\prime}=\mathbf{W}_{1} \mathbf{x}_{i}+\mathbf{W}_{2} \cdot \operatorname{mean}_{j \in \mathcal{N}(i)} \mathbf{x}_{j}$.
    
    \item For GAT, we use 3 attention heads with attention dropout of $0.1$ to stabilize training. We concatenate their linear combinations instead of aggregating, to have a output of each layer to be $3 \times hidden\_dim$.
\end{enumerate}

\subsection{Results with CNN text encoder}

The results of the proposed \ourmodel{} framework with CNN used as the text-encoder are reported in Table \ref{tab: results_cnn}. We can note similar trends in the performance although the scores are slightly lower as compared to GossipCop.

\begin{table}[ht!]
\footnotesize
\begin{centering}
    \begin{tabular}{@{}llcc@{}}
    \toprule
         & \textbf{Model} & \textbf{GossipCop} & \textbf{HealthStory} \\\midrule
        \multirow{5}{*}{\textbf{Text}} & \textsc{han}$^\dagger$ & 67.20 & - \\
         & d\textsc{efend}$^\dagger$ & 75.00 & - \\
         & \textsc{safe}$^\ddagger$ & 89.50 & - \\\cmidrule{2-4}
         & \textsc{cnn} & 66.73 &  53.81 \\
         &\textsc{r}o\textsc{bert}a & 68.55 & 57.54 \\\midrule
         & Maj. sharing baseline & 77.19 & 8.20 \\\midrule
        \multirow{8}{*}{\textbf{\textsc{\ourmodel{}}}} & \multicolumn{1}{l}{\textsc{sage}} & \underline{91.11} & 56.34\\
         & \multicolumn{1}{l}{\textsc{gcn}} & \underline{91.95} & 56.84 \\
         & \multicolumn{1}{l}{\textsc{gat}} & \underline{92.41} & 56.91 \\
         & \multicolumn{1}{l}{\textsc{r-gcn}} & $\mathbf{93.48}$ & \underline{60.45}  \\
         & \multicolumn{1}{l}{\textsc{r-gat}} & $\mathbf{93.75}$ &  \underline{61.58} \\
         & \multicolumn{1}{l}{\textsc{h}y\textsc{-gcn}} & \underline{92.34} &  \underline{59.75} \\
         & \multicolumn{1}{l}{\textsc{h}y\textsc{-gat}} & \underline{91.56} & \underline{59.89} \\\bottomrule
    \end{tabular}
    \caption{F1 scores (fake class) on \textit{GossipCop} and \textit{HealthStory} using CNN as the text encoder. ($^\dagger$) denotes results reported from \citet{defend} and ($^\ddagger$) from \citet{safe}. \textbf{Bold} figure denotes significantly better than other methods for that dataset. \underline{Underscore} figures denote significantly better scores than baselines but not significantly different from each other.}
    \label{tab: results_cnn}
\end{centering}
\end{table}

\subsection{Effect of graph sparsity and frequent users}

In Table \ref{tab: ablation_full} we report the performance of all the GNN variants of the proposed \ourmodel{} framework for different subsets of highly active users.

\begin{table}[t!]
\footnotesize
\begin{centering}
    \begin{tabular}{@{}lccc@{}}
        \toprule
        \textbf{Setting} & $\rho$ & \textbf{\ourmodel{}} & \textbf{F1} \\ \midrule
        \multirow{7}{*}{excl.\textgreater 10\%} & \multirow{7}{*}{0.78} &  \textsc{sage} &  82.14  \\
        &  & \textsc{gcn} & 81.40 \\
        &  & \textsc{gat} & 81.01 \\
        &  & \textsc{r-gcn} & 81.87 \\
        &  & \textsc{r-gat} & 82.27 \\
        &  & \textsc{h}y\textsc{-gcn} & 82.16 \\
        &  & \textsc{h}y\textsc{-gat} & 81.81 \\\midrule
        \multirow{7}{*}{excl.\textgreater 5\%} & \multirow{7}{*}{0.71} &  \textsc{sage} &  76.96  \\
        &  & \textsc{gcn} & 76.87 \\
        &  & \textsc{gat} & 77.07 \\
        &  & \textsc{r-gcn} & 77.16 \\
        &  & \textsc{r-gat} & 77.32 \\
        &  & \textsc{h}y\textsc{-gcn} & 77.13 \\
        &  & \textsc{h}y\textsc{-gat} & 77.01 \\\midrule
        \multirow{7}{*}{excl.\textgreater 1\%} & \multirow{7}{*}{0.65} & \textsc{sage} & 69.52 \\
         &  & \textsc{gcn} & 69.14  \\
         &  & \textsc{gat} & 68.67 \\
         &  & \textsc{r-gcn} & 65.89 \\
        &  & \textsc{r-gat} & 65.32 \\
        &  & \textsc{h}y\textsc{-gcn} & 71.99 \\
        &  & \textsc{h}y\textsc{-gat} & 72.05 \\\bottomrule  
    \end{tabular}
    \caption{Results of \ourmodel{} variants on varying subsets of user nodes on GossipCop. $\rho$ denotes relative graph density.}
    \label{tab: ablation_full}
\end{centering}
\end{table}

\subsection{Community Graph}
\begin{figure*}[t!]
    \centering
        \includegraphics[width=0.9\textwidth]{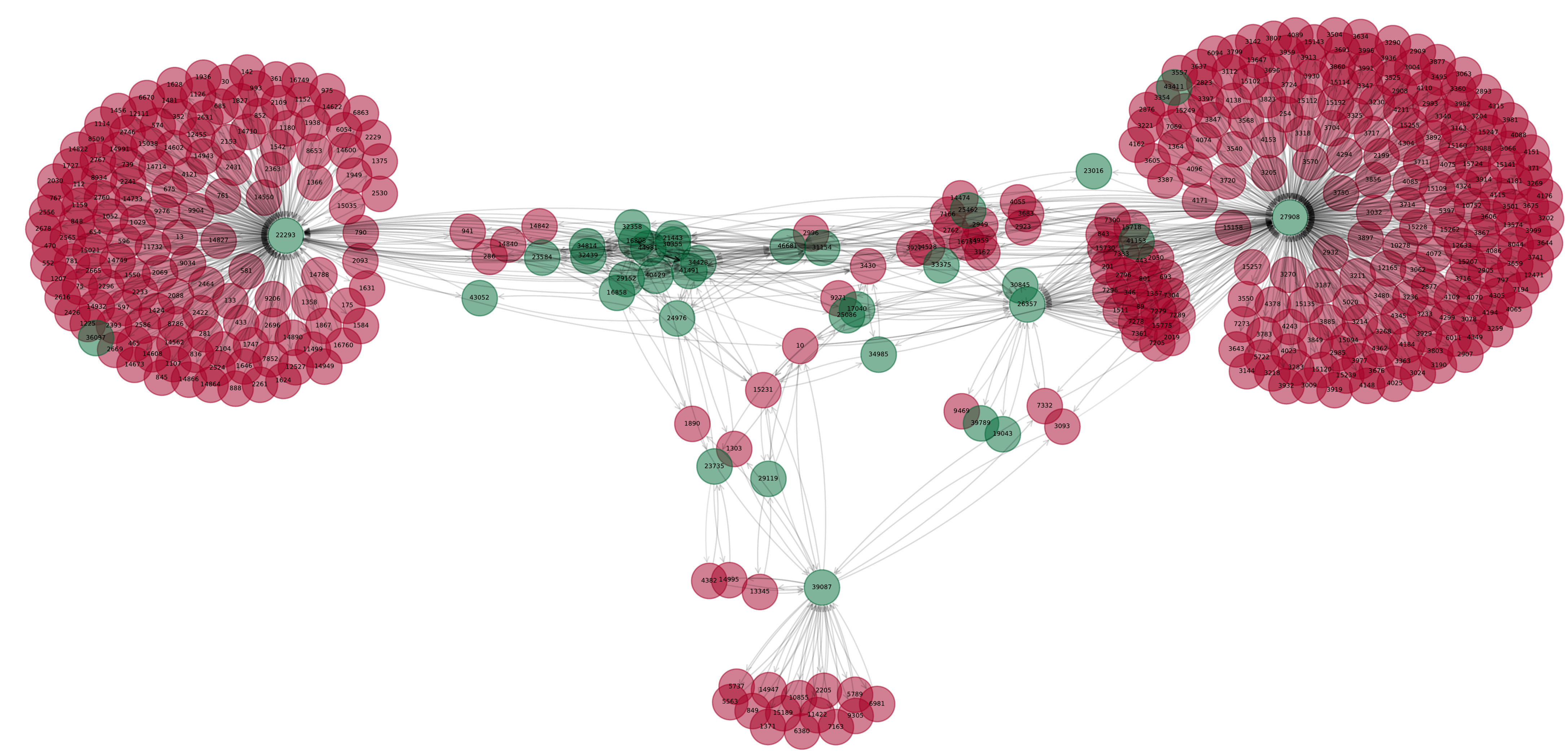}
    \caption{Visualization of a small portion of the fake news community graph. \textbf{Green} nodes represent the articles of the dataset while \textbf{red} nodes represent users that shared them.}
    \label{fig:fakenews_graph}
\end{figure*}

A portion of the community graph is visualized in Figure \ref{fig:fakenews_graph}.

\subsection{\textit{t}-SNE visualizations}

\begin{figure}[t!]
\centering
\subfloat[\label{fig:roberta}]{\includegraphics[width=0.3\textwidth]{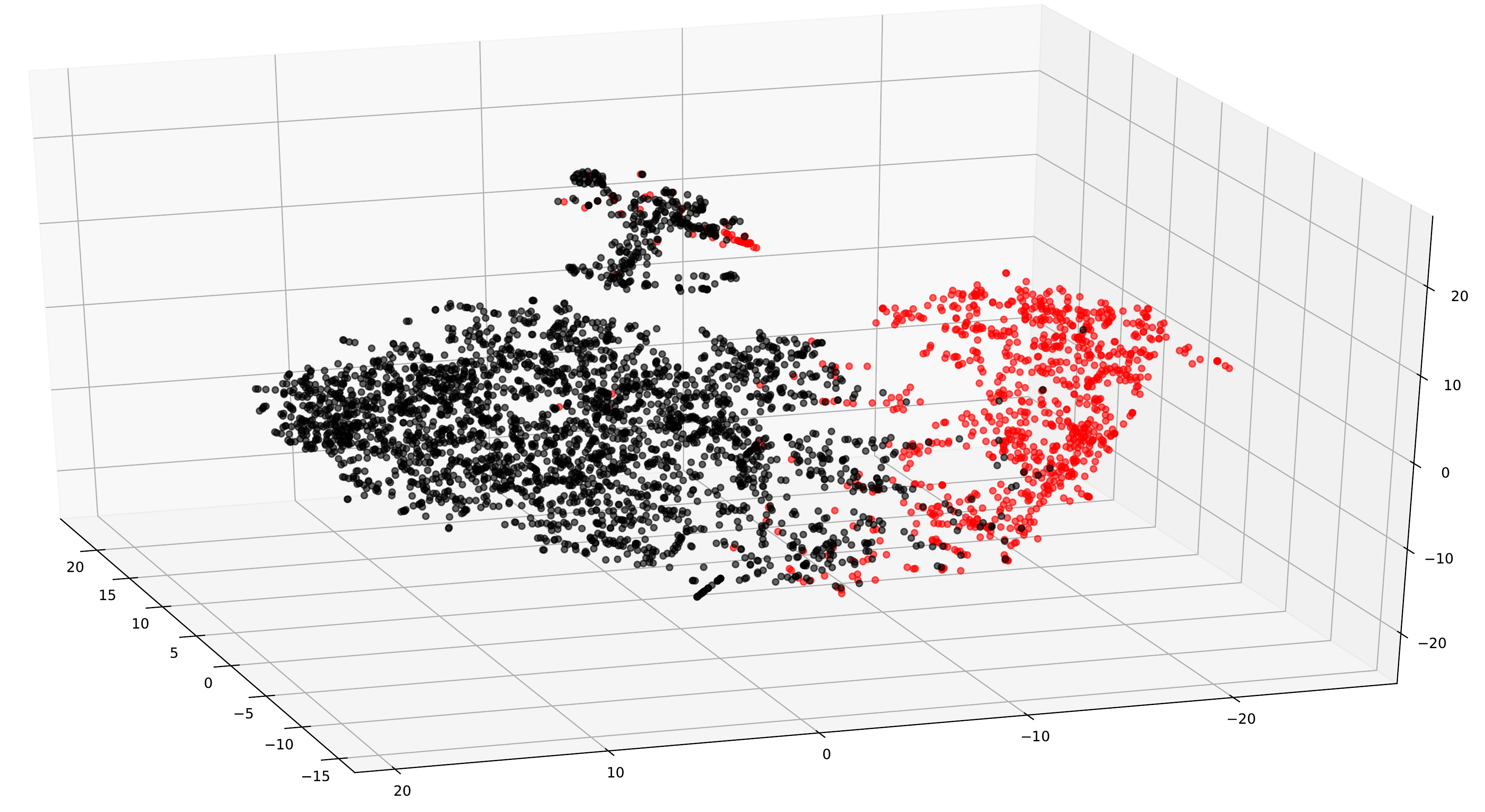}}
\quad
\subfloat[\label{fig:gat_roberta}]{\includegraphics[width=0.3\textwidth]{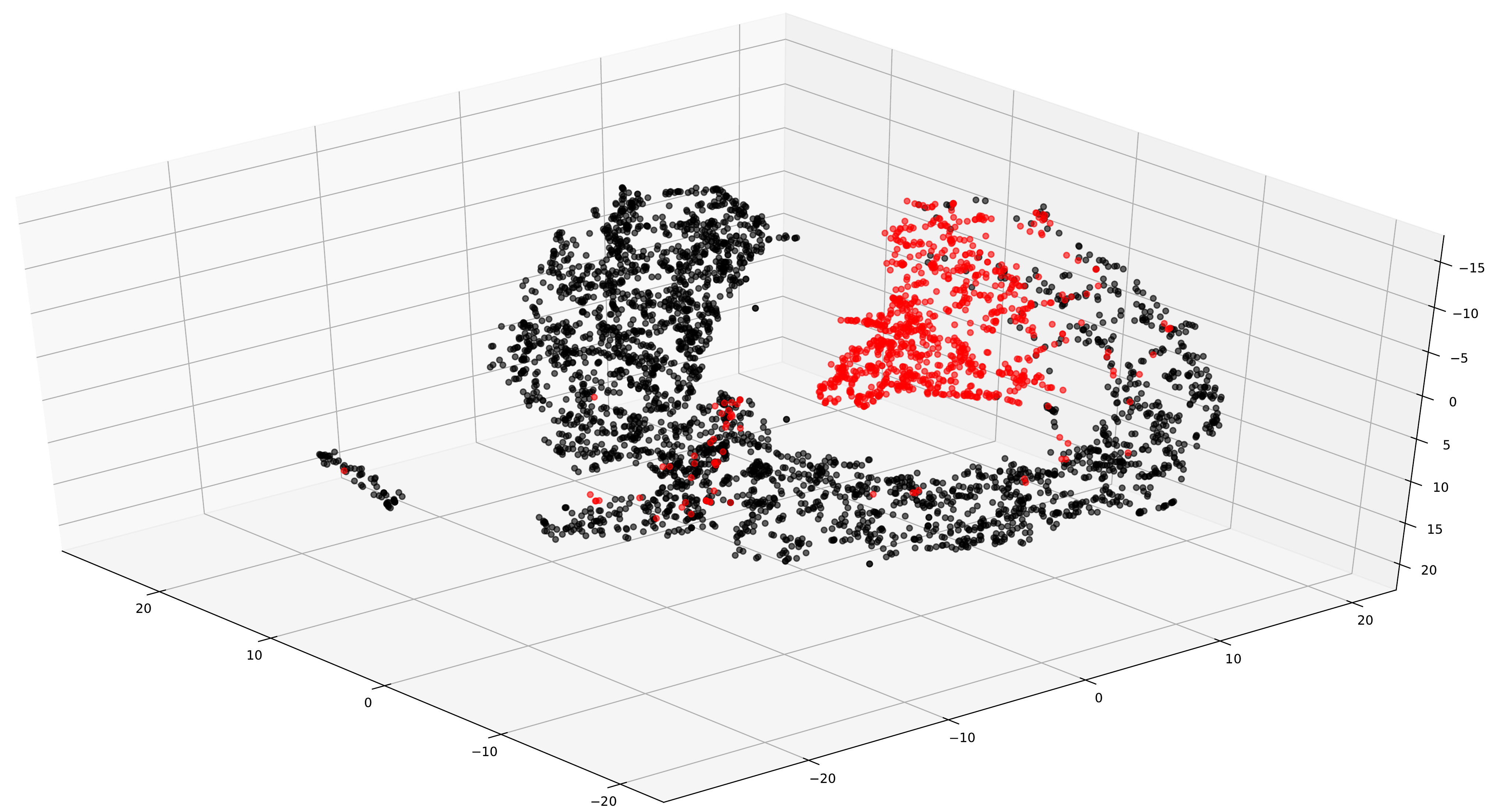}}
\quad
\subfloat[\label{fig:gcn_roberta}]{\includegraphics[width=0.3\textwidth]{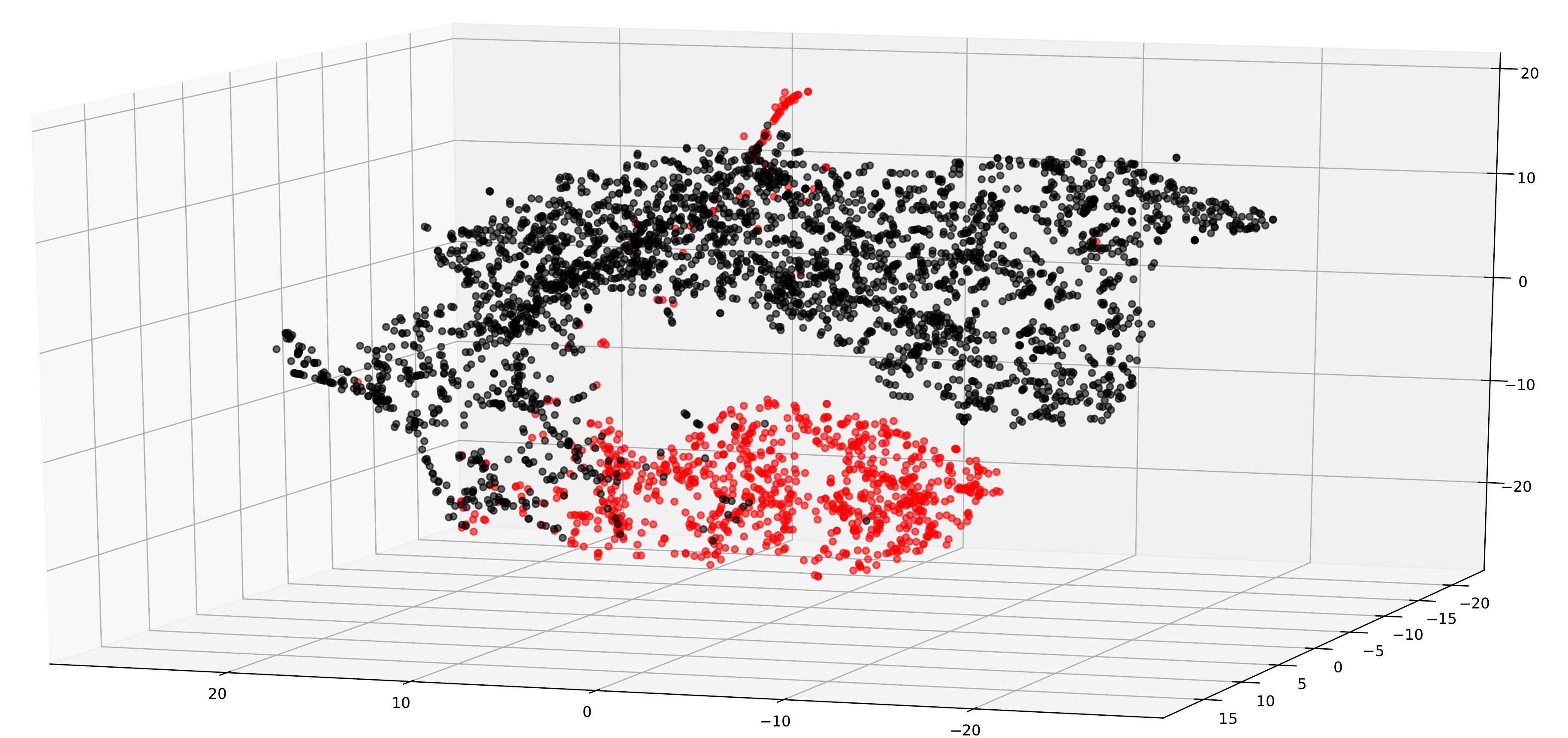}}
\caption{3-D \textit{t}-SNE plots for representations of test articles produced by (a) \ourmodel{}(GAT) (b) \ourmodel{}(GCN), and (c) \ourmodel{}(RGCN). Red dots denote fake articles.}
\end{figure}

\subsection{Qualitative Analysis}

\begin{figure*}[t!]
\centering
\subfloat[\label{fig:qual_gossip}]{\includegraphics[width=1.0\textwidth]{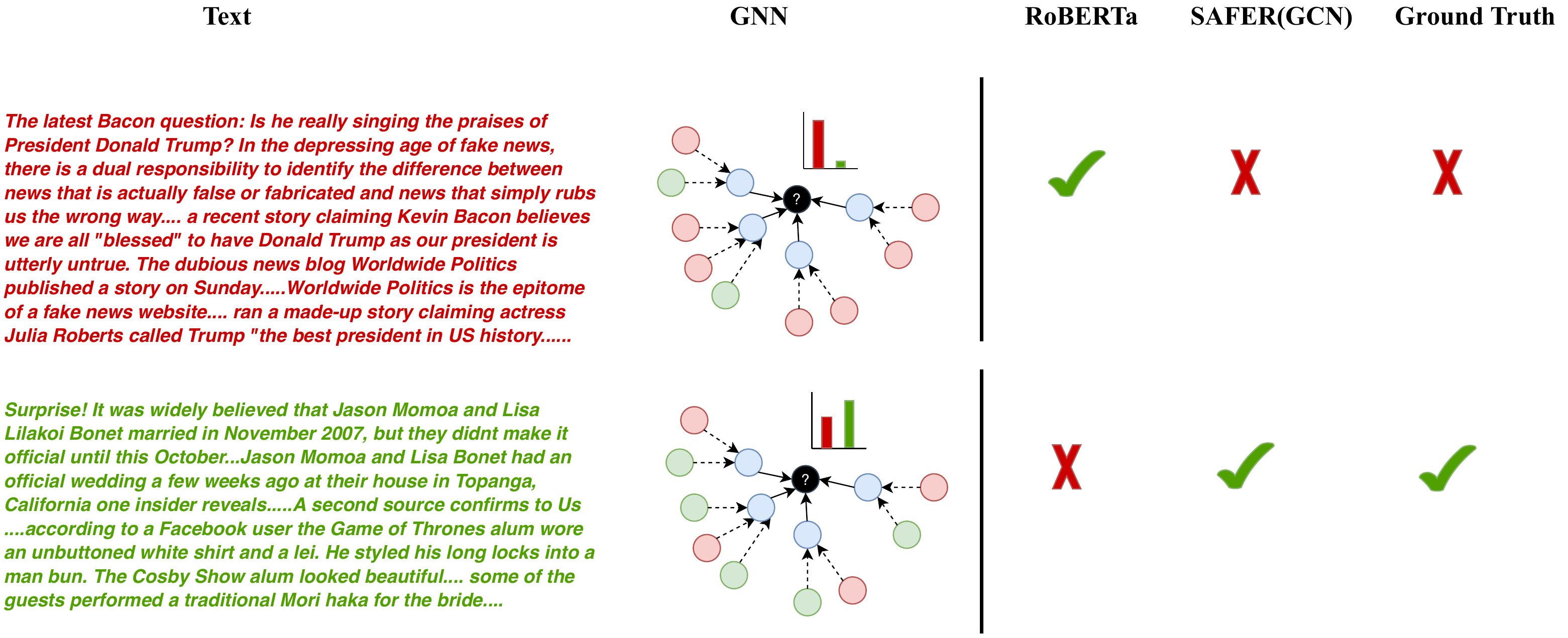}}
\vspace{5mm}
\subfloat[\label{fig:qual_health}]{\includegraphics[width=1.0\textwidth]{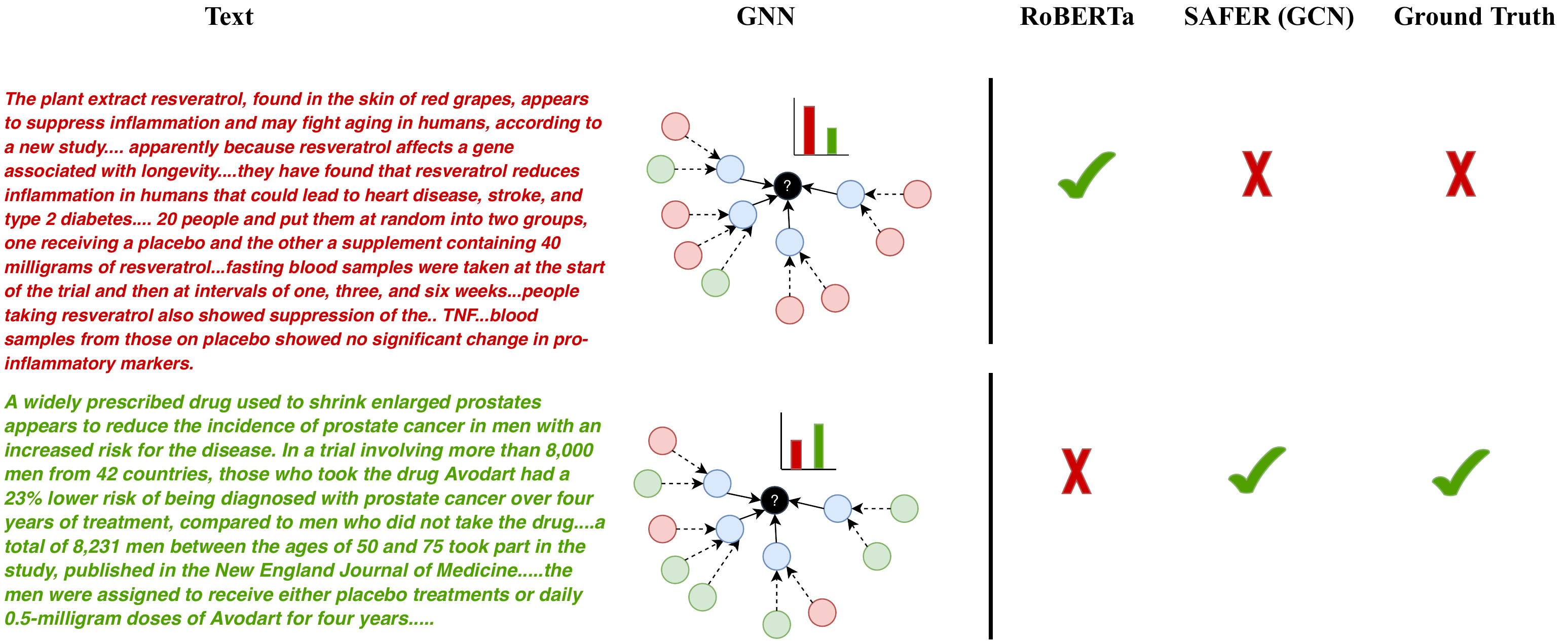}}
\caption[Qualitative analysis of GossipCop and HealthStory]{Demonstrating the importance of community-based features of the proposed method on (a) Gossipcop and, (b) HealthStory. Text in red denotes a fake article, while in green denotes a real one. Black central node denotes the target article node that we are trying to classify, blue nodes denote the users that shared this article while red and green nodes denote the other fake and real articles these users have interacted with respectively. Predictions by different models stated on the right.}
\label{fig:qual_fakenews}
\end{figure*}

We assess the performance of the \ourmodel{} (GCN) variant on Gossipcop in Figure \ref{fig:qual_gossip}. We see that the first article is a fake article which RoBERTa incorrectly classifies as real. However, looking at the content-sharing behavior of users that shared this article, we see that on average these users shared 5.8 fake articles while just 0.45 real ones (13 times more likely to share fake content than real), strongly indicating that the community of users that are involved in sharing of this article are responsible for propagation of fake news. Taking this strong community-based information into consideration, \ourmodel{} is able to correctly classify this article as fake. Similarly, the second article is a real article which is misclassified as fake by RoBERTa by looking at the text alone. However, the GNN features show that the users that shared this article have on average shared 533 real articles and 96.7 fake ones (5.5 times more likely to share a real article than a fake one). This is taken as a strong signal that the users are reliable and do not engage in malicious sharing of content. \ourmodel{} is then able to correctly classify this article as real.

We observe similar behavior of the models on HealthStory in Figure \ref{fig:qual_health}. The first article is misclassified as real by RoBERTa but the GNN features indicate that the users interacting with the article share 16.2 fake articles and 7.8 real ones on average (2.1 times more likely to share fake). \ourmodel{} takes this information into account and classifies it correctly as fake. Similarly for the second article, the interacting users share 40 real and 19.96 fake articles on average (2 times more likely to share real) which helps the proposed method to correctly classify it as real.

\end{document}